\documentclass[11pt,a4paper]{article}

\usepackage[utf8]{inputenc}
\usepackage[hyperref]{./bib_bst_sty/acl2021}
\usepackage{times}
\usepackage{latexsym}
\usepackage{xcolor}
\usepackage{url}
\usepackage{setspace}
\usepackage{xspace}
\usepackage{amsmath}
\usepackage{amsfonts}
\usepackage{booktabs}
\usepackage{subcaption}
\usepackage{graphicx}
\usepackage{mathtools}
\usepackage{soul}
\usepackage{etoolbox}
\usepackage{algorithm}
\usepackage{verbatim}

\def\aclpaperid{4272}
\aclfinalcopy

\title{H-Transformer-1D: Fast One-Dimensional Hierarchical Attention for Sequences}

\author{
  Zhenhai Zhu \\
  Google Research  \\
  \texttt{zhenhai@google.com} \\
  \And
  Radu Soricut \\
  Google Research  \\
  \texttt{rsoricut@google.com} \\
}

\date{}

\begin{document}
\maketitle

\begin{abstract} 
We describe an efficient hierarchical method to compute attention in the Transformer architecture. The proposed attention mechanism exploits a matrix structure similar to the Hierarchical Matrix (H-Matrix) developed by the numerical analysis community, and has linear run time and memory complexity. We perform extensive experiments to show that the inductive bias embodied by our hierarchical attention is effective in capturing the hierarchical structure in the sequences typical for natural language and vision tasks. Our method is superior to alternative sub-quadratic proposals by over +6 points on average on the Long Range Arena benchmark.
It also sets a new SOTA test perplexity on One-Billion Word dataset with 5x fewer model parameters than that of the previous-best Transformer-based models.
\end{abstract} 

\section{Introduction}
\label{sec:intro}
Linearly combining information using content-based weights, a method generically known as attention, is a key building block in many deep neural networks such as recurrent neural networks (RNN)~\cite{Luong2015EffectiveAT}, convolutional neural networks (CNN)~\cite{CnnAttention2019} and graph convolutional networks (GCN)~\cite{GraphAttention2018}. One particular type of such attention, called multi-head scaled dot-product attention, is one of the main components of the Transformer architecture proposed by~\citet{transformer2017}, which has been shown to push the state-of-the-art (SOTA) performance for various understanding and generation tasks. These include standard natural language processing (NLP) tasks such as machine translation, document classification, entailment, summarization and question answering~\cite{BigBird2020, TransformerXL, AdaptiveIR}, as well as music generation~\cite{MusicTransformer2018}, image generation~\cite{ImageTransformer2018, GPT-image} and genomics~\cite{BigBird2020, Performer2020}. The Transformer is also the backbone architecture for models such as BERT~\cite{BERT2019} (and its numerous relatives) and GPT3~\cite{GPT3}, which have delivered impressive performance across many NLP tasks. However, the standard attention mechanism of the Transformer has a run time and memory usage that scales quadratically with sequence length. Therefore, this quadratic complexity has become a critical bottleneck in processing long sequences (over 1,000 tokens), and has since motivated many new attention algorithms, see \cite{SurveyEfficientTransformer} for a survey of such work.

In this paper, we draw inspiration from two branches in numerical analysis: Hierarchical Matrix (H-Matrix)~\cite{hackbush99,hackbush00} and Multigrid method~\cite{multigrid_tutorial}. We propose a hierarchical attention that has linear complexity in run time and memory, and only utilizes dense linear algebra operations optimized for GPUs or TPUs. 

We hypothesize that the inductive bias embodied by the proposed hierarchical structure for the attention matrix is effective in capturing the hierarchical structure in the sequences typically seen in many natural language processing and computer vision tasks. The main benchmark we use in this paper is the Long Range Arena (LRA) benchmark~\cite{Tay2020LongRA}, which has been specifically designed to evaluate and compare various sub-quadratic attention algorithms.
Our new hierarchical attention mechanism achieves best average performance to-date on the LRA benchmark by more than 6 points over the previous-best BigBird algorithm~\cite{BigBird2020}, while pushing SOTA performance higher in 4 of the 5 successful tasks. Furthermore, using this new attention, a Transformer-based language model trained on the One-Billion Word dataset~\cite{lm1b} sets a new SOTA performance record by reducing the test perplexity by $1.55$ points comparing to the previous-best Transformer-XL~\cite{TransformerXL} with 5x more parameters. Overall, these empirical results both validate the soundness of our approximation method for computing attention weights, as well as the the appropriateness of the inductive bias present in the proposed hierarchical attention.

\section{Related Works}
It is well established in the NLP literature that the embeddings of nearby tokens tend to be more similar than the distant ones~\cite{statistical_nlp_book}. This leads to the intuition that token similarity and hence the attention should decrease with the sequence distance between a query token and a key token\footnote{Eq.~\eqref{eq:fast_decay_func} and \eqref{eq:cosine_sim} offer a simple illustration of this intuition.}. This motivates the sliding-window local attention~\cite{ImageTransformer2018, StandAloneSelfAttention2019, BlockwiseSelfAttention} which amounts to truncating off-diagonal entries in the attention matrix beyond a user-specified sequence distance.
A second approach is to keep $O(1)$ number of nonzeros per row in the attention matrix. The nonzero entry selection is either content-based~\cite{Reformer2020,RoutingTransformer,Sinkhorn2020,Informer2020}, hand-crafted~\cite{Longformer2020,GPT3,SparseTransformer2019,AxialTransformer2019} or simply random~\cite{BigBird2020}. It is also well known in the NLP literature that long-range contextual information is necessary for many NLP tasks~\cite{Khandelwal2018SharpNF, HierarchicalTransformer2019}. So a set of global tokens are also considered. This adds $O(1)$ number of dense rows and columns to the attention matrix~\cite{BigBird2020,ETC2020,Longformer2020}.
A third approach is to approximate the attention matrix with a low-rank factored form~\cite{Performer2020,Linformer2020,Synthesizer2020}. 

The first two approaches are based on the premise that one needs to explicitly zero out entries in the attention matrix in order to reduce the quadratic complexity. Decades of research by the scientific computing and numerical analysis community has resulted in more sophisticated algorithms to sparsify matrices. A small set of samples of these algorithms and their engineering applications include Fast Multipole Method~\cite{greengard87, greengard94, nabors93, shi98}, Pre-corrected FFT~\cite{phillips97, zhzhu03}, Hierarchical Singular Value Decomposition (SVD)~\cite{kapur97a} and Hierarchical Matrix (H-Matrix)~\cite{hackbush99,hackbush00, zhzhu05}. These are generally called Multilevel Methods~\cite{brandt90}.
The hierarchical attention proposed in this paper is inspired by these Multilevel Methods in general and the H-Matrix in particular. The hierarchical matrix structure allows a linear complexity in both constructing and applying the attention matrix.

\section{Definition and Notation}
\label{sec:notation}
Given matrices $Q$, $K$ and $V$, with rows representing sequences of token embedding or feature vectors for query, key and value respectively, the output weighted by the scaled dot-product attention in the Transformer~\cite{transformer2017} is defined as 
\begin{eqnarray}
  Z &=& \mathbf{softmax}(\frac{Q K^T}{\sqrt{d}}) V 
\label{eq:def_apply_attention}
\end{eqnarray}
where $Z, Q, K, V \in R^{L \times d}$, $L$ is the length of the sequences, and $d$ is the embedding or feature size. In a more compact matrix form, Eq.~\eqref{eq:def_apply_attention} can be written as 
\begin{eqnarray}
  Z = D^{-1} A V 
\label{eq:matrix_def_apply_attention}
\end{eqnarray}
where
\begin{eqnarray}
    A &=& e^{S}
\label{eq:attention_mat} \\
    S_{i,j} &=& \frac{Q_i K_j^T}{\sqrt{d}}
\label{eq:similarity_mat} \\
D &=& \mathbf{diag} \{A \cdot \mathbf{1}_L\}
\label{eq:softmax_partition} \\ 
  \mathbf{}{1}_L &=& [1, 1, ..., 1]^T.
\label{eq:all_ones} 
\end{eqnarray}
Here, $A, S \in R^{L \times L}$, $\mathbf{1}_L \in R^{L}$ is a vector with all ones,
and $S_{i, j}$ represents the unnormalized cosine similarity between query embedding $Q_i$ (the $i$-th row in $Q$) and key embedding $K_j$ (the $j$-th row in $K$). %

For the sake of clarity, we focus on the single-head attention in the exposition of the proposed algorithm. Extension to the multi-head case is straightforward since each attention head is computed independently~\cite{transformer2017}.

Computing the similarity matrix $S$ in Eq.~\eqref{eq:similarity_mat} and the attention matrix $A$ in Eq.~\eqref{eq:attention_mat} takes $O(L^2d)$ time and $O(L^2)$ memory.
Similarly, computing $AV$ in Eq.~\eqref{eq:matrix_def_apply_attention} takes $O(L^2d)$ time, and computing $A \cdot \mathbf{1}_L$ in Eq.~\eqref{eq:softmax_partition} takes $O(L^2)$ time. The $O(L^2d)$ and $O(L^2)$ complexities are the bottlenecks for applying the attention mechanism over very long sequences.

\section{Introduction on H-Matrix and Multigrid Method}
\label{sec:tutorial}

\subsection{H-Matrix}
\label{subsec:h-mat}
The singular-value decomposition of the attention matrix $A$ in Eq.~\eqref{eq:attention_mat} is
\begin{equation}
    A = U \Sigma V^T
\label{eq:svd}
\end{equation}
where $\Sigma=\mathbf{diag} \{\sigma_1,\sigma_2,...,\sigma_L\}$ and $\sigma_i$ is the $i$-th singular value. 
The numerical rank of matrix $A$ is $r$ if $\sum_{i=r+1}^{L} \sigma_i < \epsilon$ for a given tolerance $\epsilon$~\cite{trefethen_book}.
The standard rank-$r$ approximation to matrix $A$ is
\begin{equation}
    A \approx \hat{U} \hat{\Sigma} \hat{V}^T = \hat{U} \Tilde{V}^T
\label{eq:approx_svd}
\end{equation}
where $\hat{\Sigma}=\mathbf{diag} \{\sigma_1,\sigma_2,...,\sigma_r\}$, $\hat{U},\hat{V} \in R^{L \times r}$ have the first $r$ columns of $U$ and $V$, and $\Tilde{V}=\hat{V} \hat{\Sigma}$. This is the low-rank approximation used in~\cite{Performer2020,Linformer2020,Synthesizer2020}. This approximation compresses $L^2$ entries in $A$ to $2rL$ entries in $\hat{U}$ and $\Tilde{V}^T$. So the compression rate is $\frac{L}{2r}$.

The H-Matrix generalizes this low-rank approximation by using matrix block hierarchy. Consider a two-level H-Matrix with $4 \times 4$ and $2 \times 2$ block partition at level-0 and level-1, respectively. Matrix $A$ is partitioned as
\begin{equation}
    A = 
    \left[\begin{array}{@{}c|c@{}}
    \begin{array}{c|c}
       A^{(0)}_{11} &A^{(0)}_{12}\\
        \hline
       A^{(0)}_{21} &A^{(0)}_{22}
    \end{array}
    & A^{(1)}_{12} \\
    \hline
      A^{(1)}_{21} &
    \begin{array}{c|c}
       A^{(0)}_{33} &A^{(0)}_{34}\\
        \hline
       A^{(0)}_{43} &A^{(0)}_{44}
    \end{array}
    \end{array}\right].
\label{eq:h-mat}
\end{equation}
The low-rank approximation in Eq.~\eqref{eq:approx_svd} is applied to the off-diagonal blocks at each level. For example,
\begin{eqnarray}
A^{(l)}_{12} \approx 
\hat{U}^{(l)}_{12} (\Tilde{V}^{(l)}_{12})^T
  \label{eq:block_low_rank}
\end{eqnarray} 
where $l=0,1$. 
To give a concrete example, suppose each entry in matrix $A$ has the analytical form 
\begin{eqnarray}
    \label{eq:fast_decay_func}
    A_{i,j} = e^{S_{i,j}} \\
    \label{eq:cosine_sim}
    S_{i,j} = 2 e^{-(i-j)^2} - 1
\end{eqnarray}
where $i,j ={0,1,2,...,15}$~\footnote{Matrix $A$ in Eq.\eqref{eq:fast_decay_func} is a symmetric Toeplitz matrix~\cite{Golub_VanLoan_book} and hence only has 16 unique entries. But we ignore this fact and treat $A$ as a general matrix here.}. 
With the block hierarchy defined in Eq.~\eqref{eq:h-mat}, the size of the matrix block at level-1 and level-0 is $8\times 8$ and $4 \times 4$, respectively. For tolerance $\epsilon=10^{-3}$, one can verify that the numerical rank map of matrix $A$ is 
\begin{equation}
    \left[\begin{array}{@{}c|c@{}}
    \begin{array}{c|c}
        4 & 2\\
    \hline
        2 & 4
    \end{array}
    & 2 \\
    \hline
      2 &
    \begin{array}{c|c}
        4 & 2\\
    \hline
        2 & 4
    \end{array}
    \end{array}\right]
\label{eq:rank_map}
\end{equation}
where the number in each block is the numerical rank of the corresponding block in Eq.~\eqref{eq:h-mat}. Note that matrix $A$ still has full numerical rank of 16 at a looser tolerance $10^{-1}$. So the standard low-rank approximation is ineffective in this case. But even this simple two-level H-matrix already offers a compression rate of $\frac{4}{3}$ since storing an H-matrix with the rank map in Eq.~\eqref{eq:rank_map} takes $192$ entries~\footnote{Each one of four diagonal blocks at level-0 takes 16 entries. Each one of four off-diagonal blocks at level-0 takes 16 entries. Each one of two off-diagonal blocks at level-1 takes 32 entries.}. In addition, one can verify that no entry $A_{i,j}$ in Eq.~\eqref{eq:fast_decay_func} is very small, since $S_{i,j} \in [-1,1]$ in Eq.~\eqref{eq:cosine_sim}.
Therefore, truncating off-diagonal entries of matrix $A$, as proposed in~\cite{ImageTransformer2018}, would produce a poor approximation. 
In practice, the number of levels is adapted to the underlining governing equations that result in matrix $A$ and it can easily be over 10~\cite{kapur97a,hackbush00,zhzhu05}.
In turn, this can substantially increase the compression rate. 
In general, the computation complexity of the H-Matrix is either $O(L)$ or $O(L\log L)$, depending on the underlining physics~\cite{hackbush99,hackbush00}.

\subsection{Elements of the Multigrid Method}
\label{subsec:multigrid}
Multigrid Method is a multi-level nested iterative method for solving large-scale sparse matrices resulting from discretized partial-differential equations (PDEs)~\cite{multigrid_tutorial,multigrid_book}. At its core are two simple but powerfully complementary ideas: relaxation and correction. Our proposed hierarchical attention only uses the correction scheme as a building block since there is no sparse matrix to relax on.

The correction scheme has two components: restriction or coarsening, and interpolation or prolongation. Consider a vector $\bar{v}^h$ of scalar values defined on a set of $N$ grids with uniform interval $h$. The simplest coarsening is to take the average of the scalar values on each pair of grids, i.e.,
\begin{equation}
  \bar{v}^{2h}_j = \frac{1}{2} (\bar{v}^{h}_{2j} + \bar{v}^{h}_{2j+1})
\label{eq:coarsening}
\end{equation}
where $j=0,1,2,...N/2-1$. The superscript in Eq.~\eqref{eq:coarsening} indicates that the grid interval at these two levels is $h$ and $2h$, respectively. 
The simplest interpolation is to duplicate the value on each coarse grid to values on a pair of fine grids, i.e.,
\begin{equation}
  \bar{v}^{h}_{2j} = \bar{v}^{2h}_{j}, \;\;\;\;\;
  \bar{v}^{h}_{2j+1} = \bar{v}^{2h}_{j} 
\label{eq:interpolation}
\end{equation}
where $j=0,1,2,...N/2-1$. 

\section{Intuition for Hierarchical Attention}
\label{sec:intuition}

The hierarchical low-rank structure like Eq.~\eqref{eq:rank_map} turns out to be
pervasive in many if not all physics phenomena. 
Much of the theoretical analysis by~\citep{greengard87,hackbush99} is concerned with quantifying such aspects.
The key insight into these Multilevel Methods can be summarized as follows:
\emph{perform no approximation for near interactions, and apply progressively lower-precision approximation for progressively longer distance interactions}.
The simple case shown in Eq.~\eqref{eq:h-mat}-\eqref{eq:rank_map} is a good example. To satisfy the tolerance of $10^{-3}$, we need full rank (no approximation) for the diagonal blocks (near interactions), higher precision approximation (rank-2 vs full-rank of 4) for the $4 \times 4$ off-diagonal blocks at level-0 (mid-distance) and lower precision approximation (rank-2 vs full-rank of 8) for the $8 \times 8$ off-diagonal blocks at level-1 (long-distance). 

In this section, we present some intuition to answer two important questions: 1) Does the hierarchical low-rank structure hold for the attention matrix $A$ in Eq.~\eqref{eq:attention_mat}? 2) What is the algorithm to efficiently compute the hierarchical low-rank structure?
We only give an informal exposition of the hierarchical attention. The formal mathematical derivation is deferred to the Appendix.

\subsection{Hierarchical Structure As Inductive Bias}
\label{subsec:inductive_bias}
The error analysis in~\cite{greengard87,hackbush99} offers little direct insight since the attention matrix $A$ in Eq.~\eqref{eq:attention_mat} is data dependent by definition and hence its analytical form like Eq.~\eqref{eq:fast_decay_func} and ~\eqref{eq:cosine_sim} is generally unknown. 
So gathering empirical evidences seems the only viable path to answer the first question listed above.

The ablation studies by~\cite{Khandelwal2018SharpNF} examine the effect of context words on a language model. Within the context range of about 200 tokens, word order is only relevant within the 20 most recent tokens or about a sentence. In the long-range context, order has almost no effect on performance, suggesting that the model maintains a high-level, rough semantic representation of faraway words. The observation is succinctly summarized by the title of the paper "sharp nearby, fuzzy far away". Remarkably, this is in spirit very close to the key insight into the Multilevel Methods. 

A few recent attention-related studies have explored this direction with some success, such as word-level and sentence-level attentions in ~\cite{HierarchicalAttentionNMT2018, HierarchicalCNN2019}, and sentence-level and paragraph-level attentions in~\cite{HierarchicalTransformer2019}.
Even though the proposed hierarchical attention in these studies only has two levels, as opposed to ten or more levels typically used by the Multilevel Methods, the reported positive results are quite suggestive.

We therefore hypothesize that the same hierarchical low-rank structure as shown in Eq~\eqref{eq:rank_map} might also hold for the attention matrix in many NLP tasks. And we treat it as the inductive bias in the hierarchical attention mechanism proposed in this paper.
As pointed out in ~\cite{Goyal2020InductiveBF},
inductive biases encourage the learning
algorithm to prioritise solutions with certain properties. Hence good benchmark performance delivered by a Transformer-based model with proposed hierarchical attention can be regarded as a positive evidence to support the hierarchical low-rank structure hypothesis.

\subsection{Informal Exposition of Hierarchical Attention}
\label{subsec:informal_attention}
In the standard definition of attention in Eq.~\eqref{eq:attention_mat} and \eqref{eq:similarity_mat}, there is no preference given to any keys based on the sequence distance between a query and keys. The observation in~\cite{Khandelwal2018SharpNF} clearly suggests that a distance-dependent attention mechanism should be a better alternative. 

We will take three steps to informally explain the hierarchical attention mechanism.
First, the attention matrix blocks for nearby, mid-distance and long-distance attention are separated in section~\ref{subsubsec:attention_partition}. This is the first step toward the distance-dependent attention mentioned above. Second, a token hierarchy is established in section~\ref{subsubsec:token_hierarchy}. Third, the hierarchical attention is constructed in section~\ref{subsubsec:informal_construction}

\subsubsection{Attention Partition}
\label{subsubsec:attention_partition}
\begin{figure}
  \centering
  \includegraphics[width=3in]{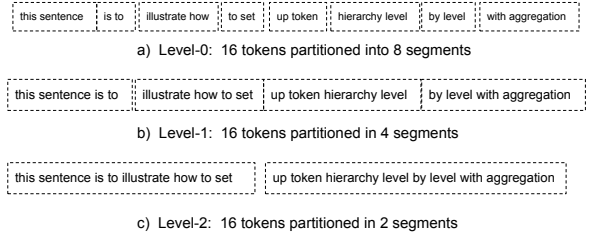}
  \caption{Token sequence partitions in three segment granularity.}
  \label{fig:token_partition}
\end{figure}

Consider a 16-word sentence in Fig.~\ref{fig:token_partition}. The sentence is partitioned at three segment granularity. This induces a three-level partition of the attention matrix $A$ for the original sequence:
\begin{equation}
    A = A^{(2)} + A^{(1)} + A^{(0)}
\label{eq:attention_partitions}
\end{equation}
where
\begin{equation}
    A^{(2)} = 
    \left[
    \begin{array}{c|c}
       0 &A^{(2)}_{12}\\
        \hline
       A^{(2)}_{21} & 0
    \end{array}
    \right]
\label{eq:attention_A2}
\end{equation}
\begin{equation}
 A^{(1)} = 
    \left[
    \begin{array}{c|c|c|c}
    & A^{(1)}_{12} & & \\
    \hline
    A^{(1)}_{21} & & A^{(1)}_{23} & \\
    \hline
    & A^{(1)}_{32} & & A^{(1)}_{34} \\
    \hline
    & & A^{(1)}_{43} &
    \end{array}
    \right]
\label{eq:attention_A1}
\end{equation}
\begin{equation}
 A^{(0)} =
    \left[
    \begin{array}{c|c|c|c|c}
    A^{(0)}_{11} &A^{(0)}_{12} & & & \\
    \hline
    A^{(0)}_{21} &A^{(0)}_{22} & A^{(0)}_{23} & & \\
    \hline
    & \ddots & \ddots & \ddots & \\
    \hline
    & & & A^{(0)}_{87} &A^{(0)}_{88}
    \end{array}
    \right].
\label{eq:attention_A0}
\end{equation}
Note that the nonzero entries in $A^{(0)}$, $A^{(1)}$ and $A^{(2)}$ are the same as the corresponding entries of matrix $A$ in Eq.~\eqref{eq:attention_mat}.
Matrix block size of $A^{(0)}_{ij}$, $A^{(1)}_{ij}$ and $A^{(2)}_{ij}$ is $2 \times 2$, $4 \times 4$ and $8 \times 8$, respectively. 
Following the key insight into Multilevel Methods, we perform no approximation to any level-0 matrix block $A^{(0)}_{ij}$ and apply a low-rank approximation to off-diagonal matrix blocks in $A^{(1)}$ and $A^{(2)}$. If we set the numerical rank of all these blocks to 2, then 
we can assemble the three rank maps into a single rank map as
~\footnote{We omit some of implementation details to handle the overlapping entries between adjacent levels.}
\begin{equation}
    \left[
    \begin{array}{@{}c|c@{}}
    \begin{array}{@{}c|c@{}}
    \begin{array}{c|c}
        2 & 2\\
    \hline
        2 & 2
    \end{array}
    & 2 \\
    \hline
      2 &
    \begin{array}{c|c}
        2 & 2\\
    \hline
        2 & 2
    \end{array}
    \end{array}
    & 2 \\
    \hline
      2 &
    \begin{array}{@{}c|c@{}}
    \begin{array}{c|c}
        2 & 2\\
    \hline
        2 & 2
    \end{array}
    & 2 \\
    \hline
      2 &
    \begin{array}{c|c}
        2 & 2\\
    \hline
        2 & 2
    \end{array}
    \end{array}
    \end{array}
    \right].
\label{eq:attention_rank_map}
\end{equation}
The hierarchical structure embodied by the predetermined rank map in Eq.~\eqref{eq:attention_rank_map} represents the inductive bias for the attention matrix $A$ in Eq.~\eqref{eq:attention_partitions}.
But this construction step is inefficient because we need to form the original attention matrix and then perform SVD to discover the low-rank approximation. 

\subsubsection{Token Hierarchy}
\label{subsubsec:token_hierarchy}
To illustrate the notion of token hierarchy, consider the same 16-word sentence in Fig.~\ref{fig:token_hierarchy}. A simple 3-level binary-tree hierarchy can be set up by following the simple coarsening defined in Eq.~\eqref{eq:coarsening}: 1)
At level-0, each one of the 16 words is mapped to its word embedding; 2) At level-1, each token (parent node) corresponds to a pair of adjacent words at level-0 (child nodes), which are shown inside each box. The embedding of each parent token is simply the average of its child token embeddings; 3) At level-2, each token (parent node) corresponds to one pair of adjacent tokens at level-1 (child nodes) or 4 adjacent words at level-0 (grand child nodes), which are shown inside each box. The embedding of each parent token is simply the average of its child token embeddings. 

In general, the height of the binary tree is $O(log_2(L)$ and the total number of tree nodes is $O(2L)$, where $L$ is the sequence length. 
We only need word embeddings for the leaf nodes since the embeddings of all other tree nodes can be recursively computed. 
The formal definition and notations of the recursion for query and key are detailed in section~\ref{subsec:constructing}.

\subsubsection{Informal Construction of Hierarchical Attention}
\label{subsubsec:informal_construction}
\begin{figure}
  \centering
  \includegraphics[width=3in]{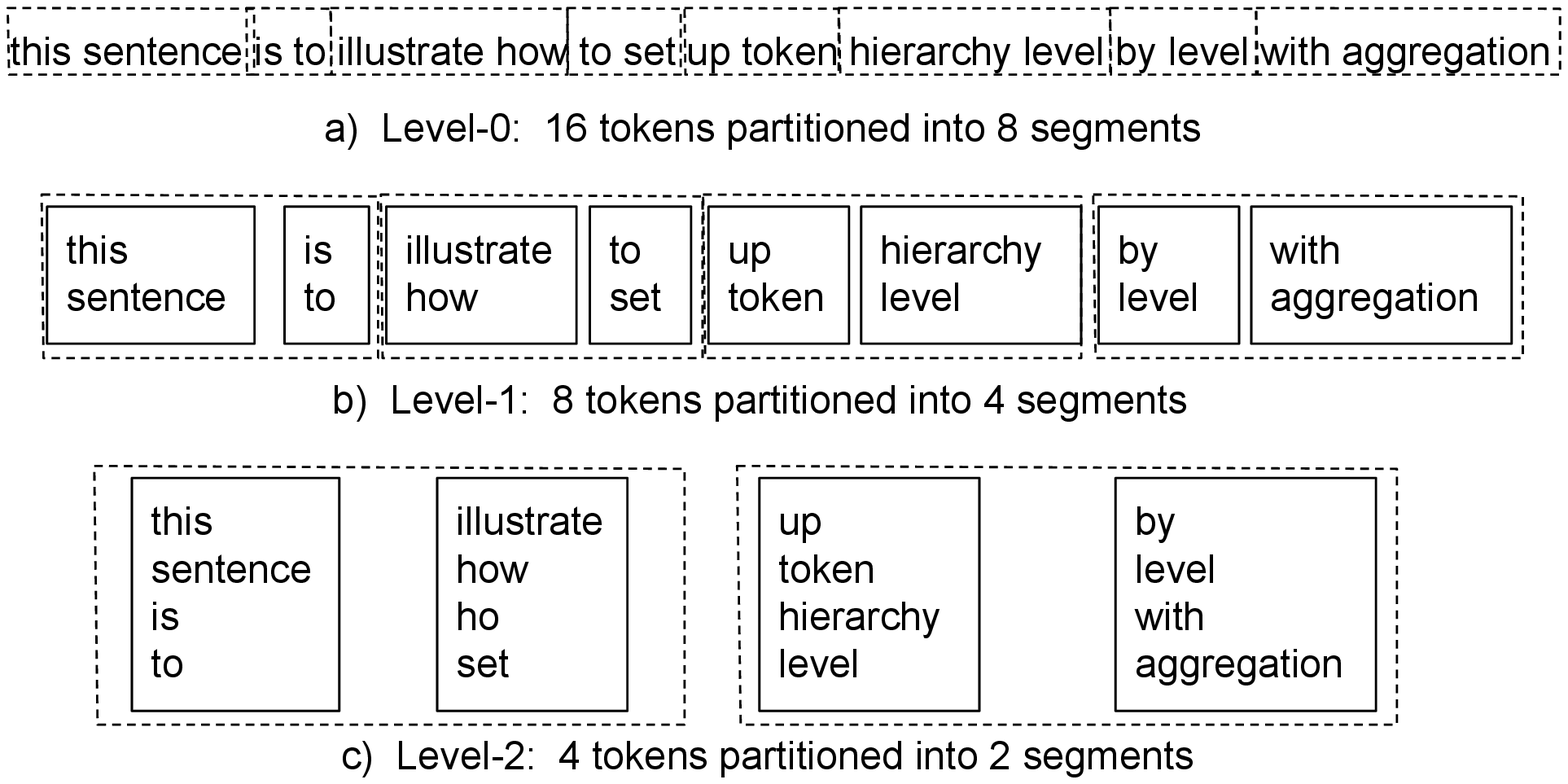}
  \caption{A three-level token hierarchy. Dashed boxes represent segmentation and solid boxes represents tokens.}
  \label{fig:token_hierarchy}
\end{figure}

It is clear from Fig.~\ref{fig:token_hierarchy} that the embeddings of higher level tokens represent a coarser level representation of a larger chunk of the text. The tokens at different levels can be understood as multi-scale snapshots of the original token sequence at level-0. Hence this token hierarchy naturally induces a set of multi-scale attention matrices. Let $\tilde{A}^{(i)}$ be the attention matrix induced by the tokens at level-$i$. It is clear from Fig.~\ref{fig:token_hierarchy} that the size of $\tilde{A}^{(0)}$, $\tilde{A}^{(1)}$ and $\tilde{A}^{(2)}$ is $16 \times 16$, $8 \times 8$ and $4 \times 4$, respectively.
This multi-scale viewpoint does not directly lead to a useful algorithm since matrix $\tilde{A}^{(0)}$ contains all the information and there is little additional information from $\tilde{A}^{(1)}$ and $\tilde{A}^{(2)}$.

A key step to arrive at the hierarchical attention is to apply the contextual sliding window at each hierarchy level. The tokens at each level are partitioned into segments of size 2 in Fig.~\ref{fig:token_hierarchy}. 
One way to implement the local attention is to allow each query token segment to attend only two adjacent key token segments, one to its left and another to its right. At level-0, each query token segment also attends to the collocated key token segment. 
The token segment partition and local attention lead to a tri-diagonal block sparse matrix structure for $\tilde{A}^{(0)}$ and bi-diagonal block sparse matrix structure for $\tilde{A}^{(1)}$ and $\tilde{A}^{(2)}$. Their sparsity patterns are
\begin{equation}
 \tilde{A}^{(0)} \propto
    \left[
    \begin{array}{c|c|c|c|c|c|c|c}
    2 & 2 & & & & & & \\
    \hline
    2 & 2 & 2 & & & & & \\
    \hline
     & 2 & 2 & 2 & & & & \\
    \hline
     & & 2 & 2 & 2 & & & \\
    \hline
     & & & 2 & 2 & 2 & & \\
    \hline
    & & & & 2 & 2 & 2 & \\
    \hline
    & & & & & 2 & 2 & 2 \\
    \hline
    & & & & & & 2 & 2
    \end{array}
    \right]
\label{eq:A0_block_structure}
\end{equation}
\begin{equation}
 \tilde{A}^{(1)} \propto 
    \left[
    \begin{array}{c|c|c|c}
    & 2 & & \\
    \hline
    2 & & 2 & \\
    \hline
    & 2 & & 2 \\
    \hline
    & & 2 &
    \end{array}
    \right]
\label{eq:A1_block_structure}
\end{equation}
\begin{equation}
 \tilde{A}^{(2)} \propto 
    \left[
    \begin{array}{c|c}
    & 2 \\
    \hline
    2 & 
    \end{array}
    \right]
\label{eq:A2_block_structure}
\end{equation}
where the 2 in the nonzero blocks indicates that these are dense blocks of size $2 \times 2$.  

It is clear that $\tilde{A}^{(0)}$ is identical to $A^{(0)}$ in Eq.~\eqref{eq:attention_A0}. 
The efficiency gain comes from $\tilde{A}^{(2)}$ and $\tilde{A}^{(1)}$.
Each nonzero entry in $\tilde{A}^{(2)}$ and $\tilde{A}^{(1)}$ captures the aggregated or coarse attention between two disjoint chunk of four and two tokens, respectively. 
Progressively larger token chunks lead to progressively lower-precision approximation to the original attention blocks. This is precisely the intention of the rank map in Eq.~\eqref{eq:attention_rank_map}.
We can now see that $\tilde{A}^{(2)}$ and $\tilde{A}^{(1)}$ provide an efficient way to approximate $A^{(2)}$ in Eq.~\eqref{eq:attention_A2} and $A^{(1)}$ in Eq.~\eqref{eq:attention_A1}, respectively.

\section{Key Components in Hierarchical Attention}
\label{sec:h_attention}

\subsection{Constructing Hierarchical Attention}
\label{subsec:constructing}
The simple example in Fig.~\ref{fig:token_hierarchy} can be easily generalized. 
Eq.~\eqref{eq:coarsening} is used to coarsen or merge rows in matrices $Q$, $K$ and $V$ in Eq.~\eqref{eq:def_apply_attention}. For sequence length $L=2^{M+1}$, the coarsening establishes a binary tree of depth $M$ for $Q$, $K$ and $V$, respectively. Each tree node represents a matrix row and there are $2^{M+1-l}$ nodes or rows at level-$l$.
To facilitate the discussion, we define a few hierarchy related notations here.
Let $\tilde{Q}^{(l)}$, $\tilde{K}^{(l)}$ and $\tilde{V}^{(l)}$ be coarsened versions of $Q$, $K$ and $V$ at level-$l$ in the binary tree. 
We note that $l=0$ is a special case, which is defined as
\begin{equation}
  \tilde{Q}^{(0)} = Q, \;\; 
  \tilde{K}^{(0)} = K, \;\;
  \tilde{V}^{(0)} = V.
\label{eq:qkv_level_0}
\end{equation}
Following Eq.~\eqref{eq:coarsening}, the recursion to coarsen $Q$, $K$ and $V$ is:
\begin{eqnarray}
\label{eq:coarsen_q}
  \tilde{Q}^{(l+1)}_j &=& \frac{1}{2} 
  (\tilde{Q}^{(l)}_{2j} + \tilde{Q}^{(l)}_{2j+1}) \\
\label{eq:coarsen_k}
  \tilde{K}^{(l+1)}_j &=& \frac{1}{2} 
  (\tilde{K}^{(l)}_{2j} + \tilde{K}^{(l)}_{2j+1}) \\
\label{eq:coarsen_v}
  \tilde{V}^{(l+1)}_j &=& 
  (\tilde{V}^{(l)}_{2j} + \tilde{V}^{(l)}_{2j+1})
\end{eqnarray}
where $l=0,1,...,M-2$ and $j=0,1,2,...,2^{M-l}$. It should be noted that the coarsening of $V$ in Eq.~\eqref{eq:coarsen_v} does not have the averaging factor $\frac{1}{2}$. 
We defer more details on coarsening to Appendix Section A.1.

Now we are ready to compute the nonzero entries in Eq.~\eqref{eq:A0_block_structure}, \eqref{eq:A1_block_structure} and \eqref{eq:A2_block_structure} and construct hierarchical attention matrix $\tilde{A}^{(l)}$. Substituting Eq.~\eqref{eq:coarsen_q} and \eqref{eq:coarsen_k} into \eqref{eq:similarity_mat} and then into \eqref{eq:attention_mat}, we obtain
\begin{equation}
    \tilde{A}^{(l)}_{ij} = e^{\tilde{S}^{(l)}_{ij}} = e^{\frac{\tilde{Q}^{(l)}_{i} (\tilde{K}^{(l)}_{j})^T}{\sqrt{d}}}
\label{eq:coarse_A_l}
\end{equation}
Again, we note that $l=0$ is a special case because $\tilde{A}^{(0)}_{ij} = A_{ij}$.

\subsection{Applying Hierarchical Attention}
\label{subsec:apply_attention}
The hierarchical matrix structure in Eq.~\eqref{eq:attention_A2}, \eqref{eq:attention_A1} and \eqref{eq:attention_A0} naturally leads to a hierarchical approach to the matrix-matrix  multiplication in Eq.~\eqref{eq:matrix_def_apply_attention} and the matrix-vector multiplication in Eq.~\eqref{eq:softmax_partition}. 
We use the matrix-matrix multiplication as an example since matrix-vector multiplication is just a special case of the matrix-matrix multiplication.

In view of Eq.~\eqref{eq:attention_A2}, \eqref{eq:attention_A1} and \eqref{eq:attention_A0}, we write the matrix-matrix multiplication in Eq.~\eqref{eq:matrix_def_apply_attention} as
\begin{eqnarray}
  Y = A V =
    Y^{(0)} + 
    P^{(0)} \left( 
        \tilde{Y}^{(1)} + P^{(1)} \tilde{Y}^{(2)}
        \right)
\label{eq:mv_mult}
\end{eqnarray}
where
\begin{equation}
  Y^{(0)} = A^{(0)} V^{(0)}, \;
  \tilde{Y}^{(l)} = \tilde{A}^{(l)} \tilde{V}^{(l)}, \; l=1,2
\label{eq:interp_mat_l}
\end{equation}
We defer the detailed derivation of Eq.~\eqref{eq:mv_mult} to Appendix Section A.5 and A.6.

\section{Algorithm And Computational Complexity}
\label{sec:algorithm} 
To facilitate the description and the complexity analysis of the algorithm, we define a few more hierarchy-related notations. In addition to sequence length $L$, number of hierarchy levels $M$ and embedding or feature size $d$ in Eq.~\eqref{eq:def_apply_attention}, the new notations include:
1) $N_r$ : numerical rank of the off-diagonal blocks (for instance, 2 in Eq.~\eqref{eq:attention_rank_map}). This is also the diagonal block size at level-0; 2) $N_b^{(l)}$: number of blocks at level-$l$. 
Note that $L$ and $d$ are usually data-dependent hyper-parameters, while $N_r$ is the only model hyper-parameter responsible for our method's inductive bias.
In turn, $N_b^{(l)}$ and $M$ are derived parameters, computed as:
\begin{eqnarray}
  N_b^{(0)} &=& \frac{L}{N_r}, \;\;
  N_b^{(l+1)} = \frac{N_b^{(l)}}{2}
\label{eq:Nb_l} \\
  M &=& \log_2(N_b^{(0)}).
\label{eq:M}
\end{eqnarray}
It is easy to verify that
\begin{equation}
  \sum_{l=0}^{M-1} N_b^{(l)} = 
  \sum_{l=0}^{M-1} \frac{N_b^{(0)}}{2^l} 
  \approx 2 N_b^{(0)}.
\label{eq:sum_num_blk}
\end{equation}

It is important to note that only the diagonal blocks at level-0 and the super-diagonal and sub-diagonal blocks at level-$l$ are needed in applying the hierarchical attention matrix. This is clearly shown in Eq.~\eqref{eq:A0_block_structure}-  \eqref{eq:A2_block_structure}.
This means that only $N_b^{(l)} - 1$ super-diagonal and sub-diagonal blocks are computed at level-$l$. This is crucial to the overall linear complexity in run time and memory.

We should also note that all matrix blocks in coarse attention matrix $\tilde{A}^{(l)}$ have the same size $N_r \times N_r$. This is due to the rank map in Eq.~\eqref{eq:attention_rank_map}. This is crucial for efficiency reason since the single-instruction-multiple-data (SIMD) programming style supported by the dense linear algebra libraries for GPU and TPU encourages uniform tensor shapes. 

We summarize the main steps to construct and apply the hierarchical attention in  Algorithm~\ref{alg:h_attention}.

\begin{algorithm} [H]
\textbf{Input: $Q$(query), $K$(key), $V$(value)} \\
\textbf{Output: $Z$} \\
\caption{H-Transformer-1D}  
\label{alg:h_attention}
  Coarsen $Q$ using Eq.~\eqref{eq:coarsen_q} and coarsen $K$ using Eq.~\eqref{eq:coarsen_k} \\
  Compute diagonal blocks in $\tilde{A}^{(0)}$ and super-diagonal and sub-diagonal blocks in $\tilde{A}^{(l)}$ using Eq.~\eqref{eq:coarse_A_l}\\
  Coarsen $V$ using Eq.~\eqref{eq:coarsen_v} \\
  Compute $Y=AV$ in Eq.~\eqref{eq:matrix_def_apply_attention} using Eq.~\eqref{eq:mv_mult} \\
  Compute $D$ in Eq.~\eqref{eq:softmax_partition} using Eq.~\eqref{eq:mv_mult} \\
  Compute $Z=D^{-1}Y$
\end{algorithm}

The computational cost for Algorithm~\ref{alg:h_attention} has two parts: \begin{enumerate}
\item Computing the hierarchical attention matrix:
  \begin{enumerate}
    \item diagonal blocks at level-$0$: $d N_r^2 N_b^{(0)}$
    \item Super- and sub-diagonal blocks at level-$l$: $4 d N_r^2 (N_b^{(l)}-1)$
    \item total: $ 5 d L N_r = O(d L)$
  \end{enumerate}
\item Computing matrix-matrix (MM) multiplication in Eq.~\eqref{eq:matrix_def_apply_attention} and matrix-vector (MV) multiplication in Eq.~\eqref{eq:softmax_partition}:
  \begin{enumerate}
    \item MM: $5 d L N_r$
    \item MV: $5 L N_r$
    \item total: $5(d+1) L N_r = O(d L)$
  \end{enumerate}
\end{enumerate}
So the overall run time complexity of the hierarchical attention algorithm is $O(d L)$. Likewise, the memory complexity can be shown to be $O(d L)$ as well. 
We defer the detailed analysis to appendix Section A.5 and A.6.

\section{Experiments And Results}
We have implemented the proposed hierarchical attention using Jax, an open source library~\footnote{https://github.com/google/jax} for automatic gradient computation and linear algebra operations on GPUs and TPUs. All numerical operations in our algorithm use the Numpy native linear algebra functions supported by Jax. 
In all our experiments in this section, we use the standard Transformer architecture described in \cite{transformer2017} as the backbone for our H-Transformer-1D model. Unless specified otherwise, the model parameters are: number of layers is 6, number of heads is 8, word embedding size is 512 and the feed-forward module (FFN) size is 2048.  
We follow the API for the standard multihead scaled dot-product attention implementation~\footnote{https://github.com/google/flax/blob/master/flax/nn} so that we can perform a simple drop-in replacement of the standard multihead attention with our hierarchical attention implementation. This allows for an easy and fair comparison.

\subsection{Long-Range Arena}
\label{subsec:lra}
The open-source Long-Range Arena (LRA)  benchmark~\footnote{https://github.com/google-research/long-range-arena} has been proposed as a standard way to probe and quantify the capabilities of various xformer (long-range Transformer) architectures~\cite{Tay2020LongRA}.
In our case, it also serves to highlight the effectiveness of the inductive bias inspired by the H-Matrix method, as well as the capability of our hierarchical attention to handle long sequences.

The LRA has several desirable qualities that made us focus on it as a primary evaluation benchmark:
\textbf{generality} (restricted to encoder-only tasks to accommodate most proposals);
\textbf{simplicity} (no pretraining, no data augmentation allowed);
\textbf{difficulty} (large headroom with existing approaches);
\textbf{long-input focus} (so that modeling improvements in this area are visible);
\textbf{diverse} (6 tasks, covering math, language, image, and spatial modeling); and
\textbf{lightweight} (so that modeling improvements are measurable independently of the ability to train and run high-capacity models).

The tasks that comprise LRA are:
\textbf{ListOps} (sequences of arithmetical expressions of lengths of up to 2K that tests the ability to reason hierarchically while handling long context);
\textbf{Text} (byte/character-level text classification at document level, which both simulates longer input sequences -- max length 4K -- and increases the difficulty level);
\textbf{Retrieval} (byte/character-level document retrieval, which simulates the ability to model document similarity as a score between two independently-encoded long input sequences -- max length 4K + 4K = 8K);
\textbf{Image} (image classification based on the CIFAR-10 dataset, where an NxN image is flattened to a sequence of length N$^2$ pixels);
\textbf{Pathfinder} (long-range spatial dependency task, with images consisting of two small circles and dash-line paths that either connect the two circles or not --  image dimensions of 32x32 for a pixel sequence of length 1,024);
\textbf{Path-X} (same as Pathfinder, but for image dimensions of 128x128 for a total pixel sequence of length 16,384). The default Transformer model parameters such as number of layers and number of heads etc are pre-determined by the benchmark configuration for each task. 

The results obtained by our H-Transformer-1D model on the LRA benchmark are given in Table~\ref{tab:lra_label}.
Overall, the H-Transformer-1D model achieves 61.41 average accuracy, a +6.4 points improvement over the previous-best average performance from BigBird~\cite{BigBird2020}.
We want to highlight ListOps, Text and Retrieval because they all involve long sequences and H-Transformer-1D model improves SOTA performance by relatively large margins. These should be strong evidences to support our hypothesis in section~\ref{subsec:inductive_bias} and validate the inductive bias due to the hierarchical attention.

\begin{table*}[t]
    \centering
    \begin{tabular}{l|cccccc|c}
    \toprule
      Model    & ListOps & Text  & Retrieval &  Image & Pathfinder & Path-X & Avg \\
         \midrule
      Chance & 10.00 & 50.00 & 50.00 & 10.00 & 50.00 & 50.00 & 44.00\\
        Transformer &  36.37 & 64.27 &
        57.46 &  42.44& 71.40 & FAIL & 54.39\\
           \midrule
        Local Attention &  15.82 &52.98 & 53.39 & 41.46& 66.63 & FAIL & 46.06 \\
        Sparse Trans.  & 17.07 & 63.58 & \underline{59.59} & \underline{44.24} & 71.71 & FAIL  & 51.24 \\
        Longformer& 35.63& 62.85 & 56.89&  42.22 & 69.71 & FAIL & 53.46 \\
        Linformer &   35.70 & 53.94&  52.27 & 38.56 & \underline{76.34} & FAIL & 51.36\\
        Reformer &  \underline{37.27} & 56.10 & 53.40 &  38.07& 68.50 & FAIL & 50.67 \\
        Sinkhorn Trans. &33.67 & 61.20 & 53.83 & 41.23 & 67.45 & FAIL & 51.39\\
        Synthesizer & 36.99 & 61.68 & 54.67  &41.61 & 69.45 & FAIL & 52.88\\
        BigBird  & 36.05 & 64.02 & 59.29  &  40.83 & 74.87 & FAIL & \underline{55.01} \\
        Linear Trans. & 16.13&  \underline{65.90} & 53.09 & 42.34 & 75.30 & FAIL & 50.55 \\
        Performer  &18.01& 65.40 & 53.82 & 42.77 & \textbf{77.05} & FAIL & 51.41\\
        H-Transformer-1D & \textbf{49.53} & \textbf{78.69} & \textbf{63.99} & \textbf{46.05} & 68.78 & FAIL & \textbf{61.41} \\
      \bottomrule
    \end{tabular}
    \caption{Experimental results on long-range arena benchmark. Best model is in boldface and second best is underlined. All models do not learn anything on Path-X task, contrary to the Pathfinder task and this is denoted by FAIL. Path-X is not counted toward the Average score as it has no impact on relative performance.}
    \label{tab:lra_label}
\end{table*}

\begin{table*}[t]
  \centering
  \begin{tabular}{l|c|c}
    \toprule
    Model & perplexity & parameters  \\
    \midrule
    \cite{TransformerXL} & 21.8 & 800M  \\
    \cite{AdaptiveIR} & 23.02 & 1000M \\
    \cite{TransformerXL} & 23.5 & 465M  \\
    \cite{AdaptiveIR} & 23.91 & 465M \\ 
    \cite{MeshTensorFlow2018} & 24.0 & 4900M \\
    \midrule
    Transformer baseline & 30.04 & 53M \\
    Transformer baseline & 24.8 & 144M \\
    H-Transformer-1D $N_r=16$ & 23.95 & 53M \\
    H-Transformer-1D $N_r=16$ & \textbf{20.25} & 144M \\
    \bottomrule
  \end{tabular}
  \caption{Experimental results on one-billion word benchmark.
  We compare previous SOTA results obtained with models of size 465M-4900M parameters against the performance of the quadratic attention baseline and the H-Transformer-1D models.
  }
  \label{tab:lm1b}
\end{table*}

\subsection{Language Models Trained on One-Billion Words}
\label{subsec:lm1b}
We have used Flax, an open-source library~\footnote{https://github.com/google/flax} to train neural networks, as the code base for the model training. Our H-Transformer-1D model uses the standard Transformer decoder implementation in Flax as the backbone. Only the attention is replaced with our hierarchical attention.  
We trained both the Transformer baseline and H-Transformer-1D on the One-Billion Word benchmark~\cite{lm1b}. 
We tried different $N_r$ (numerical rank) in our H-Transformer-1D model. These represent different inductive bias. 
We found that H-Transformer-1D with $N_r=16$ generated text with quality comparable to that of the baseline Transformer.
For both Transformer baseline and H-Transformer-1D, we also tried two sets of model parameters: 1) embedding size is 512 and feed-forward module size is 2048 and hence the parameter count is 53M; 
2) embedding size is 1024 and feed-forward module size is 4096 and hence the parameter count is 144M.
The test perplexity results of these four models and various SOTA models are shown in table~\ref{tab:lm1b}.  

H-Transformer-1D delivers the lowest perplexity to-date while using $5\times$ smaller model capacity than that of the previous SOTA model Transformer-XL~\cite{TransformerXL}. This is another strong evidence to support our hypothesis in section~\ref{subsec:inductive_bias} and validate the inductive bias due to the hierarchical attention.

\section{Conclusions and Future Work}
\label{sec:conclusion_future}
We have proposed a new Transformer attention using the inductive bias inspired by the H-Matrix. The new algorithm has linear complexity in run time and memory usage and is fully compatible with dense linear algebra libraries on GPU and TPU.
The effectiveness of this new attention is demonstrated by the empirical evidences from long-range arena benchmark and One-Billion word language modeling.
Future work include applying the new attention to music and genomics, developing proper inductive bias for cross-attention and extending to 2D images.

\bibliography{./bib_bst_sty/attention, ./bib_bst_sty/nlp, ./bib_bst_sty/numerics} 

\begin{thebibliography}{50}
\expandafter\ifx\csname natexlab\endcsname\relax\def\natexlab#1{#1}\fi

\bibitem[{Abreu et~al.(2019)Abreu, Fred, Mac{\^e}do, and
  Zanchettin}]{HierarchicalCNN2019}
Jader Abreu, Luis Fred, David Mac{\^e}do, and C.~Zanchettin. 2019.
\newblock Hierarchical attentional hybrid neural networks for document
  classification.
\newblock \emph{ArXiv}, abs/1901.06610.

\bibitem[{Ainslie et~al.(2020)Ainslie, Onta{\~n}{\'o}n, Alberti, Cvicek,
  Fisher, Pham, Ravula, Sanghai, Wang, and Yang}]{ETC2020}
Joshua Ainslie, S.~Onta{\~n}{\'o}n, C.~Alberti, V.~Cvicek, Zachary~Kenneth
  Fisher, Philip Pham, Anirudh Ravula, S.~Sanghai, Qifan Wang, and L.~Yang.
  2020.
\newblock Etc: Encoding long and structured inputs in transformers.
\newblock In \emph{EMNLP}.

\bibitem[{Baevski and Auli(2019)}]{AdaptiveIR}
Alexei Baevski and M.~Auli. 2019.
\newblock Adaptive input representations for neural language modeling.
\newblock \emph{ArXiv}, abs/1809.10853.

\bibitem[{Bello et~al.(2019)Bello, Zoph, Vaswani, Shlens, and
  Le}]{CnnAttention2019}
I.~Bello, Barret Zoph, Ashish Vaswani, Jonathon Shlens, and Quoc~V. Le. 2019.
\newblock Attention augmented convolutional networks.
\newblock \emph{2019 IEEE/CVF International Conference on Computer Vision
  (ICCV)}, pages 3285--3294.

\bibitem[{Beltagy et~al.(2020)Beltagy, Peters, and Cohan}]{Longformer2020}
Iz~Beltagy, Matthew~E. Peters, and Arman Cohan. 2020.
\newblock Longformer: The long-document transformer.
\newblock \emph{ArXiv}, abs/2004.05150.

\bibitem[{Brandt and Lubrecht(1990)}]{brandt90}
A.~Brandt and A.~A. Lubrecht. 1990.
\newblock Multilevel matrix multiplication and fast solution of integral
  equations.
\newblock 90:348--370.

\bibitem[{Briggs et~al.(2000)Briggs, Henson, and
  Mc{C}ormick}]{multigrid_tutorial}
W.L. Briggs, V.E. Henson, and S.F. Mc{C}ormick. 2000.
\newblock \emph{A Multigrid Tutorial}.
\newblock SIAM.

\bibitem[{Brown et~al.(2020)Brown, Mann, Ryder, Subbiah, Kaplan, Dhariwal,
  Neelakantan, Shyam, Sastry, Askell, Agarwal, Herbert-Voss, Kr{\"u}ger,
  Henighan, Child, Ramesh, Ziegler, Wu, Winter, Hesse, Chen, Sigler, Litwin,
  Gray, Chess, Clark, Berner, McCandlish, Radford, Sutskever, and
  Amodei}]{GPT3}
Tom~B. Brown, Benjamin~Pickman Mann, Nick Ryder, Melanie Subbiah, Jean Kaplan,
  Prafulla Dhariwal, Arvind Neelakantan, Pranav Shyam, Girish Sastry, Amanda
  Askell, Sandhini Agarwal, Ariel Herbert-Voss, G.~Kr{\"u}ger, Tom Henighan,
  Rewon Child, Aditya Ramesh, Daniel~M. Ziegler, Jeffrey Wu, Clemens Winter,
  Christopher Hesse, Mark Chen, Eric~J Sigler, Mateusz Litwin, Scott Gray,
  Benjamin Chess, Jack Clark, Christopher Berner, Sam McCandlish, Alec Radford,
  Ilya Sutskever, and Dario Amodei. 2020.
\newblock Language models are few-shot learners.
\newblock \emph{ArXiv}, abs/2005.14165.

\bibitem[{Chelba et~al.(2014)Chelba, Mikolov, Schuster, Ge, Brants, Koehn, and
  Robinson}]{lm1b}
Ciprian Chelba, Tomas Mikolov, M.~Schuster, Qi~Ge, T.~Brants, Phillipp Koehn,
  and T.~Robinson. 2014.
\newblock One billion word benchmark for measuring progress in statistical
  language modeling.
\newblock \emph{ArXiv}, abs/1312.3005.

\bibitem[{Chen et~al.(2020)Chen, Radford, Child, Wu, Jun, Luan, and
  Sutskever}]{GPT-image}
Mark Chen, Alec Radford, Rewon Child, Jeffrey Wu, Heewoo Jun, David Luan, and
  Ilya Sutskever. 2020.
\newblock Generative pretraining from pixels.
\newblock \emph{Proceedings of the 37th International Conference on Machine
  Learning}, PMLR 119.

\bibitem[{Child et~al.(2019)Child, Gray, Radford, and
  Sutskever}]{SparseTransformer2019}
R.~Child, Scott Gray, A.~Radford, and Ilya Sutskever. 2019.
\newblock Generating long sequences with sparse transformers.
\newblock \emph{ArXiv}, abs/1904.10509.

\bibitem[{Choromanski et~al.(2020)Choromanski, Likhosherstov, Dohan, Song,
  Davis, Sarl{\'o}s, Belanger, Colwell, and Weller}]{Performer2020}
Krzysztof Choromanski, Valerii Likhosherstov, David Dohan, Xingyou Song, Jared
  Davis, Tam{\'a}s Sarl{\'o}s, David Belanger, Lucy~J. Colwell, and Adrian
  Weller. 2020.
\newblock Masked language modeling for proteins via linearly scalable
  long-context transformers.
\newblock \emph{ArXiv}, abs/2006.03555.

\bibitem[{Dai et~al.(2019)Dai, Yang, Yang, Carbonell, Le, and
  Salakhutdinov}]{TransformerXL}
Zihang Dai, Z.~Yang, Yiming Yang, J.~Carbonell, Quoc~V. Le, and
  R.~Salakhutdinov. 2019.
\newblock Transformer-xl: Attentive language models beyond a fixed-length
  context.
\newblock In \emph{ACL}.

\bibitem[{Devlin et~al.(2019)Devlin, Chang, Lee, and Toutanova}]{BERT2019}
J.~Devlin, Ming-Wei Chang, Kenton Lee, and Kristina Toutanova. 2019.
\newblock Bert: Pre-training of deep bidirectional transformers for language
  understanding.
\newblock In \emph{NAACL-HLT}.

\bibitem[{Golub and Loan(1996)}]{Golub_VanLoan_book}
G.H. Golub and C.F.~Van Loan. 1996.
\newblock \emph{Matrix Computation}.
\newblock The John Hopkins University Press, Baltimore.

\bibitem[{Goyal and Bengio(2020)}]{Goyal2020InductiveBF}
Anirudh Goyal and Yoshua Bengio. 2020.
\newblock Inductive biases for deep learning of higher-level cognition.
\newblock \emph{ArXiv}, abs/2011.15091.

\bibitem[{Greengard(1994)}]{greengard94}
L~Greengard. 1994.
\newblock Fast algorithms for classical physics.
\newblock \emph{Science}, 265:909--914.

\bibitem[{Greengard and Rokhlin(1987)}]{greengard87}
L~Greengard and V~Rokhlin. 1987.
\newblock A fast algorithm for particle simulations.
\newblock 73:325--348.

\bibitem[{Hackbusch(1999)}]{hackbush99}
W.~Hackbusch. 1999.
\newblock A sparse matrix arithmetic based on h-matrices. part {I}:
  Introduction to {H}-matrices.
\newblock \emph{Computing}, 62:89--108.

\bibitem[{Hackbusch(2000)}]{hackbush00}
W.~Hackbusch. 2000.
\newblock A sparse matrix arithmetic based on {H}-matrices. part {II}:
  Application to multi-dimensional problems.
\newblock \emph{Computing}, 64:21--47.

\bibitem[{Ho et~al.(2019)Ho, Kalchbrenner, Weissenborn, and
  Salimans}]{AxialTransformer2019}
Jonathan Ho, Nal Kalchbrenner, Dirk Weissenborn, and Tim Salimans. 2019.
\newblock Axial attention in multidimensional transformers.
\newblock \emph{ArXiv}, abs/1912.12180.

\bibitem[{Huang et~al.(2018)Huang, Vaswani, Uszkoreit, Shazeer, Simon,
  Hawthorne, Dai, Hoffman, Dinculescu, and Eck}]{MusicTransformer2018}
Cheng-Zhi~Anna Huang, Ashish Vaswani, Jakob Uszkoreit, Noam Shazeer, Ian Simon,
  Curtis Hawthorne, Andrew~M. Dai, Matthew~D. Hoffman, Monica Dinculescu, and
  Douglas Eck. 2018.
\newblock Music transformer.
\newblock \emph{arXiv: Learning}.

\bibitem[{Kapur and Long(1997)}]{kapur97a}
S.~Kapur and D.E. Long. 1997.
\newblock {IES3}: A fast integral equation solver for efficient 3-dimensional
  extraction.
\newblock \emph{International Conference on Computer Aided-Design}, pages
  448--455.

\bibitem[{Khandelwal et~al.(2018)Khandelwal, He, Qi, and
  Jurafsky}]{Khandelwal2018SharpNF}
Urvashi Khandelwal, He~He, Peng Qi, and Dan Jurafsky. 2018.
\newblock Sharp nearby, fuzzy far away: How neural language models use context.
\newblock \emph{ArXiv}, abs/1805.04623.

\bibitem[{Kitaev et~al.(2020)Kitaev, Kaiser, and Levskaya}]{Reformer2020}
Nikita Kitaev, Lukasz Kaiser, and Anselm Levskaya. 2020.
\newblock Reformer: The efficient transformer.
\newblock \emph{ArXiv}, abs/2001.04451.

\bibitem[{Liu and Lapata(2019)}]{HierarchicalTransformer2019}
Yang Liu and Mirella Lapata. 2019.
\newblock Hierarchical transformers for multi-document summarization.
\newblock In \emph{ACL}.

\bibitem[{Luong et~al.(2015)Luong, Pham, and Manning}]{Luong2015EffectiveAT}
Thang Luong, Hieu Pham, and Christopher~D. Manning. 2015.
\newblock Effective approaches to attention-based neural machine translation.
\newblock \emph{ArXiv}, abs/1508.04025.

\bibitem[{Manning and Schütze(1999)}]{statistical_nlp_book}
Chris Manning and Hinrich Schütze. 1999.
\newblock \emph{Foundations of Statistical Natural Language Processing}.
\newblock MIT Press, Cambridge, MA.

\bibitem[{Miculicich et~al.(2018)Miculicich, Ram, Pappas, and
  Henderson}]{HierarchicalAttentionNMT2018}
Lesly Miculicich, Dhananjay Ram, Nikolaos Pappas, and James Henderson. 2018.
\newblock Document-level neural machine translation with hierarchical attention
  networks.
\newblock In \emph{EMNLP}.

\bibitem[{Nabors et~al.(1994)Nabors, Korsmeyer, and White}]{nabors93}
K.~Nabors, T.~Korsmeyer, and J.~White. 1994.
\newblock Multipole accelerated preconditioned iterative methods for
  three-dimensional potential integral equations of the first kind.
\newblock \emph{SIAM J. Sci. and Stat. Comp.}

\bibitem[{Parmar et~al.(2018)Parmar, Vaswani, Uszkoreit, Kaiser, Shazeer, Ku,
  and Tran}]{ImageTransformer2018}
Niki Parmar, Ashish Vaswani, Jakob Uszkoreit, Lukasz Kaiser, Noam Shazeer,
  Alexander Ku, and Dustin Tran. 2018.
\newblock Image transformer.
\newblock \emph{ArXiv}, abs/1802.05751.

\bibitem[{Phillips and White(1997)}]{phillips97}
Joel~R. Phillips and J.~K. White. 1997.
\newblock A precorrected-{FFT} method for electrostatic analysis of complicated
  {3D} structures.
\newblock \emph{{IEEE} Transactions on Computer-Aided Design of Integrated
  Circuits and Systems}, pages 1059--1072.

\bibitem[{Qiu et~al.(2019)Qiu, Ma, Levy, Yih, Wang, and
  Tang}]{BlockwiseSelfAttention}
Jiezhong Qiu, Hao Ma, Omer Levy, Scott Yih, Sinong Wang, and Jie Tang. 2019.
\newblock Blockwise self-attention for long document understanding.
\newblock \emph{ArXiv}, abs/1911.02972.

\bibitem[{Ramachandran et~al.(2019)Ramachandran, Parmar, Vaswani, Bello,
  Levskaya, and Shlens}]{StandAloneSelfAttention2019}
Prajit Ramachandran, Niki Parmar, Ashish Vaswani, Irwan Bello, Anselm Levskaya,
  and Jonathon Shlens. 2019.
\newblock Stand-alone self-attention in vision models.
\newblock \emph{ArXiv}, abs/1906.05909.

\bibitem[{Roy et~al.(2020)Roy, Saffar, Vaswani, and
  Grangier}]{RoutingTransformer}
Aurko Roy, M.~Saffar, Ashish Vaswani, and David Grangier. 2020.
\newblock Efficient content-based sparse attention with routing transformers.
\newblock \emph{ArXiv}, abs/2003.05997.

\bibitem[{Shazeer et~al.(2018)Shazeer, Cheng, Parmar, Tran, Vaswani,
  Koanantakool, Hawkins, Lee, Hong, Young, Sepassi, and
  Hechtman}]{MeshTensorFlow2018}
Noam Shazeer, Youlong Cheng, Niki Parmar, Dustin Tran, Ashish Vaswani, Penporn
  Koanantakool, P.~Hawkins, H.~Lee, Mingsheng Hong, C.~Young, Ryan Sepassi, and
  Blake~A. Hechtman. 2018.
\newblock Mesh-tensorflow: Deep learning for supercomputers.
\newblock In \emph{NeurIPS}.

\bibitem[{Shi et~al.(1998)Shi, Liu, Kakani, and Yu}]{shi98}
W.~Shi, J.~Liu, N.~Kakani, and T.~Yu. 1998.
\newblock A fast hierarchical algorithm for 3-d capacitance extraction.
\newblock \emph{{ACM}/{IEEE} Design Automation Conference}.

\bibitem[{Tay et~al.(2020{\natexlab{a}})Tay, Bahri, Metzler, Juan, Zhao, and
  Zheng}]{Synthesizer2020}
Yi~Tay, Dara Bahri, Donald Metzler, D.~Juan, Zhe Zhao, and Che Zheng.
  2020{\natexlab{a}}.
\newblock Synthesizer: Rethinking self-attention in transformer models.
\newblock \emph{ArXiv}, abs/2005.00743.

\bibitem[{Tay et~al.(2020{\natexlab{b}})Tay, Bahri, Yang, Metzler, and
  Juan}]{Sinkhorn2020}
Yi~Tay, Dara Bahri, L.~Yang, Donald Metzler, and D.~Juan. 2020{\natexlab{b}}.
\newblock Sparse sinkhorn attention.
\newblock In \emph{ICML}.

\bibitem[{Tay et~al.(2020{\natexlab{c}})Tay, Dehghani, Abnar, Shen, Bahri,
  Pham, Rao, Yang, Ruder, and Metzler}]{Tay2020LongRA}
Yi~Tay, M.~Dehghani, Samira Abnar, Y.~Shen, Dara Bahri, Philip Pham, J.~Rao,
  Liu Yang, Sebastian Ruder, and Donald Metzler. 2020{\natexlab{c}}.
\newblock Long range arena: A benchmark for efficient transformers.
\newblock \emph{ArXiv}, abs/2011.04006.

\bibitem[{Tay et~al.(2020{\natexlab{d}})Tay, Dehghani, Bahri, and
  Metzler}]{SurveyEfficientTransformer}
Yi~Tay, M.~Dehghani, Dara Bahri, and Donald Metzler. 2020{\natexlab{d}}.
\newblock Efficient transformers: A survey.
\newblock \emph{ArXiv}, abs/2009.06732.

\bibitem[{Trefethen and Bau(1997)}]{trefethen_book}
L.N. Trefethen and D.~Bau. 1997.
\newblock \emph{Numerical linear algebra}.
\newblock SIAM, Philadelphia.

\bibitem[{Trottenberg et~al.(2000)Trottenberg, Oosterlee, and
  Schuller}]{multigrid_book}
Ulrich Trottenberg, Cornelius~W. Oosterlee, and Anton Schuller. 2000.
\newblock \emph{Multigrid}.
\newblock Academic Press.

\bibitem[{Vaswani et~al.(2017)Vaswani, Shazeer, Parmar, Uszkoreit, Jones,
  Gomez, Kaiser, and Polosukhin}]{transformer2017}
Ashish Vaswani, Noam Shazeer, Niki Parmar, Jakob Uszkoreit, Llion Jones,
  Aidan~N. Gomez, Lukasz Kaiser, and Illia Polosukhin. 2017.
\newblock Attention is all you need.
\newblock \emph{ArXiv}, abs/1706.03762.

\bibitem[{Velickovic et~al.(2018)Velickovic, Cucurull, Casanova, Romero,
  Li{\`o}, and Bengio}]{GraphAttention2018}
Petar Velickovic, Guillem Cucurull, Arantxa Casanova, Adriana Romero, Pietro
  Li{\`o}, and Yoshua Bengio. 2018.
\newblock Graph attention networks.
\newblock \emph{ArXiv}, abs/1710.10903.

\bibitem[{Wang et~al.(2020)Wang, Li, Khabsa, Fang, and Ma}]{Linformer2020}
Sinong Wang, Belinda~Z. Li, Madian Khabsa, Han Fang, and Hao Ma. 2020.
\newblock Linformer: Self-attention with linear complexity.
\newblock \emph{ArXiv}, abs/2006.04768.

\bibitem[{Zaheer et~al.(2020)Zaheer, Guruganesh, Dubey, Ainslie, Alberti,
  Onta{\~n}{\'o}n, Pham, Ravula, Wang, Yang, and Ahmed}]{BigBird2020}
Manzil Zaheer, Guru Guruganesh, Kumar~Avinava Dubey, Joshua Ainslie, Chris
  Alberti, Santiago Onta{\~n}{\'o}n, Philip Pham, Anirudh Ravula, Qifan Wang,
  Li~Yang, and Amr Ahmed. 2020.
\newblock Big bird: Transformers for longer sequences.

\bibitem[{Zhou et~al.(2020)Zhou, Zhang, Peng, Zhang, Li, Xiong, and
  Zhang}]{Informer2020}
Hao-Yi Zhou, Shanghang Zhang, Jieqi Peng, Shuai Zhang, Jianxin Li, Hui Xiong,
  and Wancai Zhang. 2020.
\newblock Informer: Beyond efficient transformer for long sequence time-series
  forecasting.
\newblock \emph{ArXiv}, abs/2012.07436.

\bibitem[{Zhu et~al.(2005)Zhu, Song, and White}]{zhzhu03}
Zhenhai Zhu, Ben Song, and J.~K. White. 2005.
\newblock Algorithms in {F}ast{I}mp: A fast and wideband impedance extraction
  program for complicated 3{D} geometries.
\newblock \emph{{IEEE} Transactions on Computer-Aided Design of Integrated
  Circuits and Systems}.

\bibitem[{Zhu and White(2005)}]{zhzhu05}
Zhenhai Zhu and J.~K. White. 2005.
\newblock Fastsies: a fast stochastic integral equation solver for modeling the
  rough surface effect.
\newblock \emph{International Conference on Computer Aided-Design}, pages
  675--682.

\end{thebibliography}

\clearpage
\appendix

\section{Appendix}
\label{sec:appendix}

\subsection{Restriction or Coarsening Matrices}
\label{app_subsec:coarsen_mat}
For sequence length $L=2^{M+1}$, the coarsening establishes a binary tree of depth $M$ for $Q$, $K$ and $V$, respectively. 
The root of the binary tree at level-$(M-1)$ has two nodes which correspond to the two matrix rows coarsened from four matrix rows at level-$(M-2)$.
The piecewise constant restriction matrix at level-$(M-2)$ is 
\begin{equation}
    R^{(M-2)} = 
    \left[ 
      \begin{array}{cccc}
       1 & 1 & 0 & 0 \\
       0 & 0 & 1 & 1 
    \end{array}
    \right]_{2 \times 4}.
\label{app_eq:R_M_2}
\end{equation}
Likewise, the piecewise constant restriction matrix at level-$(M-3)$ is 
\begin{eqnarray}
    R^{(M-3)} &=& 
    \left[ 
      \begin{array}{cccc|cccc}
       1 & 1 & 0 & 0 & 0 & 0 & 0 & 0 \\
       0 & 0 & 1 & 1 & 0 & 0 & 0 & 0 \\ 
       \hline
       0 & 0 & 0 & 0 & 1 & 1 & 0 & 0 \\
       0 & 0 & 0 & 0 & 0 & 0 & 1 & 1 
    \end{array}
    \right]_{4 \times 8}
    \nonumber \\
    &=&
    \left[ 
      \begin{array}{c|c}
       R^{(M-2)} & 0 \\
       \hline
       0 & R^{(M-2)}
    \end{array}
    \right].
\label{app_eq:R_M_3}
\end{eqnarray}
In general, the restriction matrices follow the recursion
\begin{equation}
    R^{(l-1)} = 
    \left[ 
      \begin{array}{c|c}
       R^{(l)} & 0 \\
       \hline
       0 & R^{(l)}
    \end{array}
    \right]
\label{app_eq:Rl_recursion}
\end{equation}
which starts from $R^{(M-2)}$ of size $2 \times 4$ and goes backward to $R^{(0)}$ of size $\frac{L}{2} \times L$.

\subsection{Interpolation Matrices}
\label{app_subsec:interp_mat}
Given $Y^{(l)}$ at level-$l$, the interpolated $Y^{(l-1)}$ at level-$(l-1)$ can be written as
\begin{equation}
    Y^{(l-1)} = P^{(l)} Y^{(l)}
\label{app_eq:interp_recursion}
\end{equation}
where $l=1,2,...,M-1$, sparse matrix $P^{(l)}$ has size $L^{(l-1)} \times L^{(l)}$, and $L^{(l)}=2^{M-l}$ is the node count at level-$l$ of the binary tree.

This recursion also follows the binary tree hierarchy.
The four matrix rows at level-$(M-2)$ are interpolated from the two matrix rows at level-$(M-1)$.
Specifically, the piecewise constant interpolation matrix at level-$(M-1)$ is
\begin{equation}
    P^{(M-1)} = 
    \left[ 
      \begin{array}{cc}
       1 & 0 \\
       1 & 0 \\
       0 & 1 \\
       0 & 1 
    \end{array}
    \right]_{4 \times 2}.
\label{app_eq:P_M_1}
\end{equation}
Likewise, the piecewise constant interpolation matrix at level-$(M-2)$ is
\begin{eqnarray}
    P^{(M-2)} &=& 
    \left[ 
      \begin{array}{cc|cc}
       1 & 0 & 0 & 0 \\
       1 & 0 & 0 & 0 \\
       0 & 1 & 0 & 0 \\
       0 & 1 & 0 & 0 \\
       \hline
       0 & 0 & 1 & 0 \\
       0 & 0 & 1 & 0 \\
       0 & 0 & 0 & 1 \\
       0 & 0 & 0 & 1
    \end{array}
    \right]_{8 \times 4}
    \nonumber \\
    &=&
    \left[ 
      \begin{array}{c|c}
       P^{(M-1)} & 0 \\
       \hline
       0 & P^{(M-1)}
    \end{array}
    \right].
\label{app_eq:P_M_2}
\end{eqnarray}
In general, the interpolation matrices follow the recursion
\begin{equation}
    P^{(l-1)} = 
    \left[ 
      \begin{array}{c|c}
       P^{(l)} & 0 \\
       \hline
       0 & P^{(l)}
    \end{array}
    \right]
\label{app_eq:Pl_recursion}
\end{equation}
which starts from $P^{(M-1)}$ of size $4 \times 2$ and goes backward to $P^{(0)}$ of size $L \times \frac{L}{2}$.
In view of Eq.~\eqref{app_eq:R_M_2} and \eqref{app_eq:P_M_1}, it is obvious that 
\begin{equation}
  P^{(M-1)} = (R^{(M-2)})^T.
\label{app_eq:transpose_P_R_M_1}
\end{equation}
In view of the recursions in Eq.~\eqref{app_eq:Rl_recursion} and \eqref{app_eq:Pl_recursion}, it is easy to prove by induction that
\begin{equation}
  P^{(l)} = (R^{(l-1)})^T.
\label{app_eq:transpose_P_R_l}
\end{equation}

\subsection{Expansion Matrices}
\label{app_subsec:expansion_mat}
For the purpose of factored low-rank approximation for the off-diagonal attention matrix blocks, we design a series of so-called expansion matrices. The first two expansion matrices in this series are 
\begin{eqnarray}
  T^{(M-1)} &=& P^{(M-1)} =
      \left[ 
      \begin{array}{cc}
       1 & 0 \\
       1 & 0 \\
       0 & 1 \\
       0 & 1 
    \end{array}
    \right]_{4 \times 2}
    \nonumber \\
    &=&
      \left[ 
        \begin{array}{cc}
        \mathbf{1}_2 & 0 \\
        0 & \mathbf{1}_2
        \end{array}
      \right]
\label{app_eq:T_M_1}
\end{eqnarray}
and
\begin{eqnarray}
  T^{(M-2)} 
  &=& P^{(M-2)} P^{(M-1)}  =
    \left[ 
      \begin{array}{c|c}
       1 & 0  \\
       1 & 0 \\
       1 & 0 \\
       1 & 0 \\
       \hline
       0 & 1 \\
       0 & 1 \\
       0 & 1 \\
       0 & 1
    \end{array}
    \right]_{8 \times 2}
  \nonumber \\ 
  &=&
  \left[ 
  \begin{array}{cc}
       \mathbf{1}_4 & 0 \\
       0 & \mathbf{1}_4
    \end{array}
    \right]
\label{app_eq:T_M_2}
\end{eqnarray}
where $\mathbf{1}_N$ is a length-$N$ vector of ones. The general form of matrix $T^{(l)}$ is defined as
\begin{equation}
  T^{(l)} 
  = \Pi_{i=l}^{M-1} P^{(i)}
\label{app_eq:Tl_def}
\end{equation}
where $l=1,2,...,M-1$. 
In view of Eq.~\eqref{app_eq:T_M_1}, \eqref{app_eq:Tl_def} and \eqref{app_eq:Pl_recursion}, it is easy to prove by induction that
\begin{equation}
  T^{(l)} =   
  \left[ 
  \begin{array}{cc}
       \mathbf{1}_{2^{M-l}} & 0 \\
       0 & \mathbf{1}_{2^{M-l}}
    \end{array}
    \right]
\label{app_eq:Tl_form}
\end{equation}
and it has size $2^{M-l+1} \times 2$.
Further more, in view of Eq.~\eqref{app_eq:Tl_def} and \eqref{app_eq:transpose_P_R_l}, we have 
\begin{equation}
  (T^{(l)})^T 
  = \Pi_{i=M-1}^{l} R^{(i-1)}.
\label{app_eq:transpose_Tl_def}
\end{equation}

\subsection{Low-Rank Factored Form}
\label{app_subsec:factored_form}
Matrix $T^{(l)}$ plays a pivotal role in constructing the low-rank approximation to the off-diagonal attention matrix blocks.
Let the $ij$-th block in the coarsened attention matrix at level-1 be
\begin{equation}
\tilde{A}^{(1)}_{ij} = 
    \left[ 
      \begin{array}{cc}
       a_{11} & a_{12} \\
       a_{21} & a_{22}
    \end{array}
    \right]
\label{app_eq:coarsened_block}
\end{equation}
where $a_{ij}$ is the entry resulted from the inner product between a row in $\tilde{Q}^{(1)}$ and $\tilde{K}^{(1)}$.
The rank-2 approximation to the corresponding $ij$-th block in the original attention matrix $A$ at level-$1$ can be written as
\begin{eqnarray}
    &&A^{(1)}_{ij}
    \approx
    T^{(M-1)}
    \tilde{A}^{(1)}_{ij}
    (T^{(M-1)})^T
\label{app_eq:low_rank_approx} \\
    &=&
    \left[ 
      \begin{array}{cc}
       1 & 0 \\
       1 & 0 \\
       0 & 1 \\
       0 & 1 
      \end{array}
    \right]
    \left[ 
      \begin{array}{cc}
       a_{11} & a_{12} \\
       a_{21} & a_{22}
    \end{array}
    \right]
    \left[ 
      \begin{array}{cccc}
       1 & 1 & 0 & 0 \\
       0 & 0 & 1 & 1 
      \end{array}
    \right]
    \nonumber \\
    &=&
    \left[ 
      \begin{array}{cc|cc}
       a_{11} & a_{11} & a_{12} & a_{12} \\
       a_{11} & a_{11} & a_{12} & a_{12} \\
        \hline
       a_{21} & a_{21} & a_{22} & a_{22} \\
       a_{21} & a_{21} & a_{22} & a_{22}
    \end{array}
    \right].
\label{app_eq:expanded_block}
\end{eqnarray}
It is clear that the resulting $4 \times 4$ matrix $A^{(1)}_{ij}$ is essentially the piecewise constant interpolation of the $2 \times 2$ matrix $\tilde{A}^{(1)}_{ij}$ along row and column direction. And since both $T^{(M-1)}$ and $\tilde{A}^{(1)}_{ij}$ have full rank 2, $A^{(1)}_{ij}$ necessarily has rank 2. 
One can also view $a_{ij}$ as being similar to the average value at the $ij$-th cluster center in the K-mean method. The role of matrix $T^{(M-1)}$ is to expand from these $2 \times 2$ clusters to the $4 \times 4$ grid and hence the name expansion matrix.

Since we maintain the same numerical rank 2 for all super- and sub-diagonal attention matrix blocks, the rank-2 approximation to the $ij$-th block in the original attention matrix $A$ at level-$l$ is
\begin{eqnarray}
    A^{(l)}_{ij}
    &\approx&
    T^{(M-l)}
    \tilde{A}^{(l)}_{ij}
    (T^{(M-l)})^T
    \nonumber \\
    &=&
    \Pi_{i=M-l}^{M-1} P^{(i)}
    \tilde{A}^{(l)}_{ij}
    \Pi_{i=M-1}^{M-l} R^{(i-1)}
\label{app_eq:low_rank_approx_recursion}
\end{eqnarray}
where the last equality is due to Eq.~\eqref{app_eq:Tl_def} and \eqref{app_eq:transpose_Tl_def}. 

We note that matrix $T^{(l)}$ has full column rank 2 by design and this can be easily shown from Eq.~\eqref{app_eq:Tl_form}. 
We have used this fact to construct the rank-2 approximation in Eq.~\eqref{app_eq:low_rank_approx_recursion}.

\subsection{Construct Hierarchical Attention Matrix}
\label{app_subsec:construct_attention}
To see how Eq.~\eqref{app_eq:low_rank_approx_recursion} can be used, consider a simple three-level partition of the attention matrix $A$ for sequence length $L=16$
\begin{equation}
    A = 
    \left[
    \begin{array}{c|c}
       A^{(2)}_{11} &A^{(2)}_{12}\\
        \hline
       A^{(2)}_{21} &A^{(2)}_{22}
    \end{array}
    \right]
\label{app_eq:attention_partition_2}
\end{equation}
\begin{equation}
    A^{(2)}_{11} = 
    \left[\begin{array}{@{}c|c@{}}
    \begin{array}{c|c}
       A^{(0)}_{11} &A^{(0)}_{12}\\
        \hline
       A^{(0)}_{21} &A^{(0)}_{22}
    \end{array}
    & A^{(1)}_{12} \\
    \hline
      A^{(1)}_{21} &
    \begin{array}{c|c}
       A^{(0)}_{33} &A^{(0)}_{34}\\
        \hline
       A^{(0)}_{43} &A^{(0)}_{44}
    \end{array}
    \end{array}\right]
\label{app_eq:attention_partition_2_11}
\end{equation}
\begin{equation}
    A^{(2)}_{22} = 
    \left[\begin{array}{@{}c|c@{}}
    \begin{array}{c|c}
       A^{(0)}_{55} &A^{(0)}_{56}\\
        \hline
       A^{(0)}_{65} &A^{(0)}_{66}
    \end{array}
    & A^{(1)}_{34} \\
    \hline
      A^{(1)}_{43} &
    \begin{array}{c|c}
       A^{(0)}_{77} &A^{(0)}_{78}\\
        \hline
       A^{(0)}_{87} &A^{(0)}_{88}
    \end{array}
    \end{array}\right]
\label{app_eq:attention_partition_2_22}
\end{equation}
where the size of level-0, level-1 and level-2 matrix blocks is $2 \times 2$, $4 \times 4$ and $8 \times 8$, respectively. Note that the number of levels is $M=log_2(L/2)=3$. 
We use this simple three-level example to illustrate the key steps in both constructing and applying the hierarchical attention matrix.

In view of Eq.~\eqref{app_eq:low_rank_approx_recursion}, we have
\begin{equation}
    A \approx 
    \left[
    \begin{array}{c|c}
       \tilde{A}^{(2)}_{11} & 
       T^{(1)} \tilde{A}^{(2)}_{12} (T^{(1)} )^T
       \\ \hline
       T^{(1)} \tilde{A}^{(2)}_{21} (T^{(1)} )^T &
       \tilde{A}^{(2)}_{22}
    \end{array}
    \right] 
\label{app_eq:low_rank_A_partition_2}
\end{equation}
\begin{equation}
    \tilde{A}^{(2)}_{11} = 
    \left[\begin{array}{@{}c|c@{}}
    \begin{array}{c|c}
       A^{(0)}_{11} & A^{(0)}_{12}
       \\ \hline
       A^{(0)}_{21} & A^{(0)}_{22}
    \end{array}
    & T^{(2)} \tilde{A}^{(1)}_{12} (T^{(2)} )^T
    \\  \hline
    T^{(2)} \tilde{A}^{(1)}_{21} (T^{(2)} )^T &
    \begin{array}{c|c}
       A^{(0)}_{33} &A^{(0)}_{34}\\
        \hline
       A^{(0)}_{43} &A^{(0)}_{44}
    \end{array}
    \end{array}\right]
\label{app_eq:low_rank_A_partition_2_11}
\end{equation}
\begin{equation}
    \tilde{A}^{(2)}_{22} = 
    \left[\begin{array}{@{}c|c@{}}
    \begin{array}{c|c}
       A^{(0)}_{55} &A^{(0)}_{56}\\
        \hline
       A^{(0)}_{65} &A^{(0)}_{66}
    \end{array}
    & T^{(2)} \tilde{A}^{(1)}_{34} (T^{(2)} )^T
    \\  \hline
    T^{(2)} \tilde{A}^{(1)}_{43} (T^{(2)} )^T &
    \begin{array}{c|c}
       A^{(0)}_{77} &A^{(0)}_{78}\\
        \hline
       A^{(0)}_{87} &A^{(0)}_{88}
    \end{array}
    \end{array}\right].
\label{app_eq:low_rank_A_partition_2_22}
\end{equation}
We note that matrices $T^{(l)}, l=1,2$ are never explicitly formed and are only implicitly used, as shown in next section. So only the diagonal blocks at level-0 and super- and sub-diagonal blocks of the coarsened matrix $\tilde{A}$ at level-$l$ need to be explicitly computed. By design, all these blocks have the same size $2 \times 2$ if we set the numerical rank to $N_r=2$. The total number of super- and sub-diagonal blocks in the binary tree hierarchy is upper bounded by twice the number of super- and sub-diagonal blocks at level-0, which is $2N_b^{(0)}$. Hence the total number of entries is $5 N_b^{(0)} N_r^2 = 5 L N_r = O(L N_r)$. Each entry is equal to the inner product between $\tilde{Q}^{(l)}_i$ and $\tilde{K}^{(l)}_j$ and hence the run time cost per entry is $O(d)$, where $d$ is the embedding size. So the final total run time cost is $O(L d)$ and memory foot print is $O(L)$. Here we leave out $N_r$ since it is a constant model hyper parameter.

\subsection{Apply Hierarchical Attention Matrix}
\label{app_subsec:apply_attention}
Computing matrix-matrix product $AV$ follows the hierarchical structure of matrix $A$ in Eq.~\eqref{app_eq:low_rank_A_partition_2},
\eqref{app_eq:low_rank_A_partition_2_11} and \eqref{app_eq:low_rank_A_partition_2_22}. We first partition matrix $V$ according to the three-level binary tree established by the coarsening process, i.e.,
\begin{equation}
    V =
    \left[
    \begin{array}{c}
    V^{(0)}_1 \\
    V^{(0)}_2 \\
    \hline
    \vdots \\
    \hline
    V^{(0)}_7 \\
    V^{(0)}_8
    \end{array}
    \right] 
    = 
    \left[
    \begin{array}{c}
    V^{(1)}_1 \\
    V^{(1)}_2 \\
    \hline
    V^{(1)}_3 \\
    V^{(1)}_4
    \end{array}
    \right] 
    = 
    \left[
    \begin{array}{c}
    V^{(2)}_1 \\
    V^{(2)}_2
    \end{array}
    \right].
\label{app_eq:V_partition}
\end{equation}
Note that these are partitions of the same matrix $V$ at 3 different levels. For sequence length $L=16$, matrix $V$ has size $16 \times d$, and the size of the partitioned blocks $V^{(0)}_i$, $V^{(1)}_j$ and $V^{(2)}_k$ are $2 \times d$, $4 \times d$ and $8 \times d$, respectively.
In the derivation to come, we may exchange partitions at different levels. For instance, in view of Eq.~\eqref{app_eq:V_partition}, we have
\begin{equation}
    V^{(2)}_1
    = 
    \left[
    \begin{array}{c}
    V^{(1)}_1 \\
    V^{(1)}_2
    \end{array}
    \right].
\label{app_eq:V1_1_partition}
\end{equation}
So we may replace $V^{(2)}_1$ with the right-hand side in Eq.~\eqref{app_eq:V1_1_partition}.

In view of Eq.~\eqref{app_eq:attention_partition_2} and \eqref{app_eq:V_partition}, matrix-matrix product $AV$ can be written as
\begin{eqnarray}
    Y &=& AV =
    \left[
    \begin{array}{c}
       A^{(2)}_{11} V^{(2)}_1 \\
       A^{(2)}_{22} V^{(2)}_2
    \end{array}
    \right]
    +
    \left[
    \begin{array}{c}
       A^{(2)}_{12} V^{(2)}_2 \\
       A^{(2)}_{21} V^{(2)}_1
    \end{array}
    \right]
    \nonumber \\
    &=&
    \left[
    \begin{array}{c}
       A^{(2)}_{11} V^{(2)}_1 \\
       A^{(2)}_{22} V^{(2)}_2
    \end{array}
    \right]
    + Y^{(2)}.
\label{app_eq:AV_partition}
\end{eqnarray}
In view of Eq.~\eqref{app_eq:low_rank_A_partition_2}, we have 
\begin{eqnarray}
    Y^{(2)} &=&
    \left[
    \begin{array}{c}
       A^{(2)}_{12} V^{(2)}_2 \\
       A^{(2)}_{21} V^{(2)}_1
    \end{array}
    \right]
    \nonumber \\
    &\approx& 
    \left[
    \begin{array}{c}
       T^{(1)} \tilde{A}^{(2)}_{12} (T^{(1)} )^T  V^{(2)}_2
       \\
       T^{(1)} \tilde{A}^{(2)}_{21} (T^{(1)} )^T V^{(2)}_1
    \end{array}
    \right] 
    \nonumber \\
    &=&
    \left[
    \begin{array}{c}
       P^{(1)} P^{(2)} \tilde{A}^{(2)}_{12} R^{(1)} R^{(0)} V^{(2)}_2
       \\ 
       P^{(1)} P^{(2)} \tilde{A}^{(2)}_{21} R^{(1)} R^{(0)} V^{(2)}_1
    \end{array}
    \right] 
    \nonumber \\
    &=&
    P^{(0)} P^{(1)}
    \left[
    \begin{array}{c}
       \tilde{A}^{(2)}_{12} \tilde{V}^{(2)}_2 \\
       \tilde{A}^{(2)}_{21} \tilde{V}^{(2)}_1
    \end{array}
    \right]
    \nonumber \\
    &=&
    P^{(0)} P^{(1)} 
    \left[
    \begin{array}{c}
      \tilde{Y}^{(2)}_1 \\
      \tilde{Y}^{(2)}_2
    \end{array}
    \right]
\label{app_eq:Y2}
\end{eqnarray}
where
\begin{equation}
    \left[
    \begin{array}{c}
       \tilde{V}^{(2)}_1 \\
       \tilde{V}^{(2)}_2
    \end{array}
    \right]
    =
    \left[
    \begin{array}{c}
      R^{(1)} R^{(0)} V^{(2)}_1
       \\ 
      R^{(1)} R^{(0)} V^{(2)}_2
    \end{array}
    \right]. 
\label{app_eq:tilde_V2}
\end{equation}
The third equality in Eq.~\eqref{app_eq:Y2} is due to Eq.~\eqref{app_eq:Tl_def} and \eqref{app_eq:transpose_Tl_def} where $l=1$.
The fourth equality in Eq.~\eqref{app_eq:Y2} is due to Eq.~\eqref{app_eq:Pl_recursion}. 

In view of Eq.~\eqref{app_eq:low_rank_A_partition_2_11}, we have 
\begin{eqnarray}
    &&A^{(2)}_{11} V^{(2)}_1 
    \approx \tilde{A}^{(2)}_{11} V^{(2)}_1
    \nonumber \\
    &=& 
    \left[\begin{array}{@{}c|c@{}}
    \begin{array}{c|c}
       A^{(0)}_{11} & A^{(0)}_{12}
       \\ \hline
       A^{(0)}_{21} & A^{(0)}_{22}
    \end{array}
    & T^{(2)} \tilde{A}^{(1)}_{12} (T^{(2)} )^T
    \\  \hline
    T^{(2)} \tilde{A}^{(1)}_{21} (T^{(2)} )^T &
    \begin{array}{c|c}
       A^{(0)}_{33} &A^{(0)}_{34}\\
        \hline
       A^{(0)}_{43} &A^{(0)}_{44}
    \end{array}
    \end{array}\right] 
     V^{(2)}_1
    \nonumber \\
    &=&
    \left[
    \begin{array}{c}
      Y^{(0)}_1
       \\ 
      Y^{(0)}_2
       \\ 
      Y^{(0)}_3
       \\ 
      Y^{(0)}_4
    \end{array}
    \right]
     + Y^{(1)}_1
\label{app_eq:AV_2_11}
\end{eqnarray}
where
\begin{eqnarray}
    Y^{(1)}_1 
    &=&
    \left[
    \begin{array}{c}
       T^{(2)} \tilde{A}^{(1)}_{12} (T^{(2)} )^T  V^{(1)}_2
       \\
       T^{(2)} \tilde{A}^{(1)}_{21} (T^{(2)} )^T V^{(1)}_1
    \end{array}
    \right] 
    \nonumber \\
    &=&
    \left[
    \begin{array}{c}
       P^{(2)} \tilde{A}^{(1)}_{12} R^{(1)} V^{(1)}_2
       \\ 
       P^{(2)} \tilde{A}^{(1)}_{21} R^{(1)} V^{(1)}_1
    \end{array}
    \right] 
    \nonumber \\
    &=&
    P^{(1)}
    \left[
    \begin{array}{c}
       \tilde{A}^{(1)}_{12} \tilde{V}^{(1)}_2 \\
       \tilde{A}^{(1)}_{21} \tilde{V}^{(1)}_1
    \end{array}
    \right]
    \nonumber \\
    &=&
    P^{(1)} 
    \left[
    \begin{array}{c}
       \tilde{Y}^{(1)}_1 \\
       \tilde{Y}^{(1)}_2
    \end{array}
    \right]
\label{app_eq:Y1_1}
\end{eqnarray}
and
\begin{equation}
    \left[
    \begin{array}{c}
       \tilde{V}^{(1)}_1 \\
       \tilde{V}^{(1)}_2
    \end{array}
    \right]
  =
    \left[
    \begin{array}{c}
      R^{(1)} V^{(1)}_1
       \\ 
      R^{(1)} V^{(1)}_2
    \end{array}
    \right]. 
\label{app_eq:tilde_V1_1_2}
\end{equation}
The second equality in Eq.~\eqref{app_eq:Y1_1} is due to Eq.~\eqref{app_eq:Tl_def} and \eqref{app_eq:transpose_Tl_def} where $l=2$.
The third equality in Eq.~\eqref{app_eq:Y1_1} is due to Eq.~\eqref{app_eq:Pl_recursion}. 

In view of Eq.\eqref{app_eq:low_rank_A_partition_2_22}, we have 
\begin{eqnarray}
    &&A^{(2)}_{22} V^{(2)}_2 
    \approx \tilde{A}^{(2)}_{22} V^{(2)}_2
    \nonumber \\
    &=& 
    \left[\begin{array}{@{}c|c@{}}
    \begin{array}{c|c}
       A^{(0)}_{55} &A^{(0)}_{56}\\
        \hline
       A^{(0)}_{65} &A^{(0)}_{66}
    \end{array}
    & T^{(1)} \tilde{A}^{(1)}_{34} (T^{(1)} )^T
    \\  \hline
    T^{(1)} \tilde{A}^{(1)}_{43} (T^{(1)} )^T &
    \begin{array}{c|c}
       A^{(0)}_{77} &A^{(0)}_{78}\\
        \hline
       A^{(0)}_{87} &A^{(0)}_{88}
    \end{array}
    \end{array}\right] 
      V^{(2)}_2
    \nonumber \\
    &=&
    \left[
    \begin{array}{c}
      Y^{(0)}_5
       \\ 
      Y^{(0)}_6
       \\ 
      Y^{(0)}_7
       \\ 
      Y^{(0)}_8
    \end{array}
    \right]
     + Y^{(1)}_2
\label{app_eq:AV_2_22}
\end{eqnarray}
where
\begin{eqnarray}
    Y^{(1)}_2 
    &=&
    \left[
    \begin{array}{c}
       P^{(2)} \tilde{A}^{(1)}_{34} R^{(1)} V^{(1)}_4
       \\ 
       P^{(2)} \tilde{A}^{(1)}_{43} R^{(1)} V^{(1)}_3
    \end{array}
    \right] 
    \nonumber \\
    &=&
    P^{(1)}
    \left[
    \begin{array}{c}
       \tilde{A}^{(1)}_{34} \tilde{V}^{(1)}_4 \\
       \tilde{A}^{(1)}_{43} \tilde{V}^{(1)}_3
    \end{array}
    \right]
    \nonumber \\
    &=&
    P^{(1)} 
    \left[
    \begin{array}{c}
       \tilde{Y}^{(1)}_3 \\
       \tilde{Y}^{(1)}_4
    \end{array}
    \right]
\label{app_eq:Y1_2}
\end{eqnarray}
and
\begin{equation}
    \left[
    \begin{array}{c}
       \tilde{V}^{(1)}_3 \\
       \tilde{V}^{(1)}_4
    \end{array}
    \right]
    =
    \left[
    \begin{array}{c}
      R^{(1)} V^{(1)}_3
       \\ 
      R^{(1)} V^{(1)}_4
    \end{array}
    \right]. 
\label{app_eq:tilde_V1_3_4}
\end{equation}

Substituting Eq.~\eqref{app_eq:Y2}, \eqref{app_eq:AV_2_11} and \eqref{app_eq:AV_2_22} into \eqref{app_eq:AV_partition}, we obtain the final result for the matrix-matrix product
\begin{eqnarray}
    Y = AV \approx
    Y^{(0)} + 
    P^{(0)} \left( 
        \tilde{Y}^{(1)} + P^{(1)} \tilde{Y}^{(2)}
        \right)
\label{app_eq:interp_cumsum}
\end{eqnarray}
where
\begin{eqnarray}
    Y^{(0)} &=& 
    \left[
    \begin{array}{c}
       A^{(0)}_{11} V^{(0)}_1 + A^{(0)}_{12} V^{(0)}_2 \\
       A^{(0)}_{21} V^{(0)}_1 + A^{(0)}_{22} V^{(0)}_2 \\
        \vdots \\
       A^{(0)}_{87} V^{(0)}_7 + A^{(0)}_{88} V^{(0)}_8
    \end{array}
    \right]
\label{app_eq:Y0} \\
    \tilde{Y}^{(1)} &=& 
    \left[
    \begin{array}{c}
        \tilde{Y}^{(1)}_1 \\
        \tilde{Y}^{(1)}_2 \\
        \tilde{Y}^{(1)}_3 \\
        \tilde{Y}^{(1)}_4
    \end{array}
    \right]
    =    
    \left[
    \begin{array}{c}
       \tilde{A}^{(1)}_{12} \tilde{V}^{(1)}_2 \\
       \tilde{A}^{(1)}_{21} \tilde{V}^{(1)}_1 \\
       \tilde{A}^{(1)}_{34} \tilde{V}^{(1)}_4 \\
       \tilde{A}^{(1)}_{43} \tilde{V}^{(1)}_3
    \end{array}
    \right]
\label{app_eq:tilde_Y1} \\
    \tilde{Y}^{(2)} &=& 
    \left[
    \begin{array}{c}
      \tilde{Y}^{(2)}_1 \\
      \tilde{Y}^{(2)}_2
    \end{array}
    \right]
     =
    \left[
    \begin{array}{c}
       \tilde{A}^{(2)}_{12} \tilde{V}^{(2)}_2 \\
       \tilde{A}^{(2)}_{21} \tilde{V}^{(2)}_1
    \end{array}
    \right]
\label{app_eq:tilde_Y2}
\end{eqnarray}

To summarize, matrix-matrix product computation includes the following steps:
\begin{enumerate}
\item Compute $\tilde{V}^{(1)}$ in Eq.~\eqref{app_eq:tilde_V1_1_2} and  \eqref{app_eq:tilde_V1_3_4}, and compute $\tilde{V}^{(2)}$ in Eq.~\eqref{app_eq:tilde_V2};
\item Compute $Y^{(0)}$ in Eq.~\eqref{app_eq:Y0}, $\tilde{Y}^{(1)}$ in Eq.~\eqref{app_eq:tilde_Y1} and $\tilde{Y}^{(2)}$ in Eq.~\eqref{app_eq:tilde_Y2};
\item Interpolate and cumulative sum in Eq.~\eqref{app_eq:interp_cumsum};
\end{enumerate}
Note that all operations in step-2 are dense matrix-matrix product, well suited for dense linear algebra libraries optimized for GPU and TPU. 
The total number of super- and sub-diagonal blocks is upper bounded by twice the number of super- and sub-diagonal blocks at level-0, which is $2N_b^{(0)}$. The run time of each dense matrix-matrix product is $O(N_r^2 d)$. So the total run time is $5 N_b^{(0)} N_r^2 d = 5 L N_r d = O(L d)$. Here we leave out $N_r$ since it is a constant model hyper-parameter.

The coarsening in step-1 and interpolation in step-3 all use sparse matrices with fixed sparsity patterns. Hence matrices $P^{(l)}$ and $R^{(l)}$ are never explicitly formed and applying them can be easily done with standard library functions. Take Jax Numpy library as an example, coarsening can be done with sum() along row axis and interpolation can be done with repeat() along row axis. For this reason, step-1 and step-3 only have dense matrix operations as well.

The formulation of the matrix-matrix product for the general level-$M$ case is
\begin{eqnarray}
    Y &=& AV =
    Y^{(0)} + 
    P^{(0)} (\tilde{Y}^{(1)} + 
        P^{(1)} ( \tilde{Y}^{(2)}
    \nonumber \\
        &+& P^{(2)} ( \cdots + P^{(M-2)} \tilde{Y}^{(M-1)}
        ) \cdots ) ).
\label{app_eq:interp_cumsum_general}
\end{eqnarray}
This formulation is a direct consequence of the nested attention matrix structure and can be derived similarly as Eq.~\eqref{app_eq:interp_cumsum}.

\bibliographystyle{./bib_bst_sty/acl_natbib.bst}

\end{document}